\documentclass[sigconf]{acmart}
\AtBeginDocument{%
\providecommand\BibTeX{{%
\normalfont B\kern-0.5em{\scshape i\kern-0.25em b}\kern-0.8em\TeX}}}

\usepackage{graphicx}
\usepackage{amsmath}
\usepackage{amssymb}
\usepackage{balance}
\usepackage{bm}
\usepackage{subfig}
\usepackage{algorithm}
\usepackage[noend]{algpseudocode}
\usepackage{booktabs,multirow,array}
\usepackage{float}
\usepackage{color,soul}
\usepackage{xcolor}
\usepackage{caption}
\usepackage[capitalize]{cleveref}
\crefdefaultlabelformat{#2\textcolor{red}{#1}#3}
\crefname{algorithm}{Algorithm}{Algorithms}
\crefname{section}{Sec.}{Secs.}
\Crefname{section}{Section}{Sections}
\crefname{table}{Tab.}{Tabs.}
\Crefname{table}{Table}{Tables}
\crefname{figure}{Fig.}{Figs.}
\Crefname{figure}{Fig.}{Figs.}
\crefname{equation}{Eq.}{Eqs.}
\Crefname{equation}{Eq.}{Eqs}
\soulregister\cite7
\soulregister\ref7
\soulregister\pageref7
\definecolor{b}{RGB}{66,133,244}
\definecolor{r}{RGB}{234,67,53}
\definecolor{p}{RGB}{153,0,204}
\definecolor{y}{RGB}{251,188,3}
\newcommand{\etal}{{\it et al.}}
\newcommand{\eg}{{\it e.g.}}

\usepackage{tabularx}
\newcolumntype{Y}{>{\centering\arraybackslash}X}

\AtBeginDocument{%
\providecommand\BibTeX{{%
		Bib\TeX}}}

\copyrightyear{2023}
\acmYear{2023}
\setcopyright{acmlicensed}\acmConference[MM '23]{Proceedings of the 31st ACM International Conference on Multimedia}{October 29-November 3, 2023}{Ottawa, ON, Canada}
\acmBooktitle{Proceedings of the 31st ACM International Conference on Multimedia (MM '23), October 29-November 3, 2023, Ottawa, ON, Canada}
\acmPrice{15.00}
\acmDOI{10.1145/3581783.3611884}
\acmISBN{979-8-4007-0108-5/23/10}

\begin{document}

\title{Enhancing Visibility in Nighttime Haze Images Using Guided APSF and Gradient Adaptive Convolution}

\author{Yeying Jin}
\authornote{Both authors contributed equally to this research.}
\affiliation{
	\institution{National University of Singapore}
	\city{Singapore}
	\country{Singapore}}
\email{e0178303@u.nus.edu}
\orcid{0000-0001-7818-9534}

\author{Beibei Lin}
\authornotemark[1]
\affiliation{
	\institution{National University of Singapore}
	\city{Singapore}
	\country{Singapore}}
\email{beibei.lin@u.nus.edu}

\author{Wending Yan}
\affiliation{
	\institution{Huawei International Pte Ltd}
	\city{Singapore}
	\country{Singapore}}
\email{yan.wending@huawei.com}

\author{Yuan Yuan}
\affiliation{
	\institution{Huawei International Pte Ltd}
	\city{Singapore}
	\country{Singapore}}
\email{yuanyuan10@huawei.com}

\author{Wei Ye}
\affiliation{
	\institution{Huawei International Pte Ltd}
	\city{Singapore}
	\country{Singapore}}
\email{yewei10@huawei.com}

\author{Robby T. Tan}
\affiliation{
	\institution{National University of Singapore}
	\city{Singapore}
	\country{Singapore}}
\email{robby.tan@nus.edu.sg}

\renewcommand{\shortauthors}{Yeying Jin et al.}

\begin{abstract}
Visibility in hazy nighttime scenes is frequently reduced by multiple factors, including low light, intense glow, light scattering, and the presence of multicolored light sources. Existing nighttime dehazing methods often struggle with handling glow or low-light conditions, resulting in either excessively dark visuals or unsuppressed glow outputs. In this paper, we enhance the visibility from a single nighttime haze image by suppressing glow and enhancing low-light regions. To handle glow effects, our framework learns from the rendered glow pairs. Specifically, a light source aware network is proposed to detect light sources of night images, followed by the APSF (Atmospheric Point Spread Function)-guided glow rendering. Our framework is then trained on the rendered images, resulting in glow suppression. Moreover, we utilize gradient-adaptive convolution, to capture edges and textures in hazy scenes. By leveraging extracted edges and textures, we enhance the contrast of the scene without losing important structural details. To boost low-light intensity, our network learns an attention map, then adjusted by gamma correction. This attention has high values on low-light regions and low values on haze and glow regions. Extensive evaluation on real nighttime haze images, demonstrates the effectiveness of our method. Our experiments demonstrate that our method achieves a PSNR of 30.38dB, outperforming state-of-the-art methods by 13$\%$ on GTA5 nighttime haze dataset. Our data and code is available at: \url{https://github.com/jinyeying/nighttime_dehaze}.
\end{abstract}



\begin{teaserfigure}
	\centering
	\captionsetup[subfloat]{farskip=1.5pt}
	\captionsetup[subfigure]{labelformat=empty}
	\subfloat[Night Haze]{\includegraphics[width=0.245\textwidth]{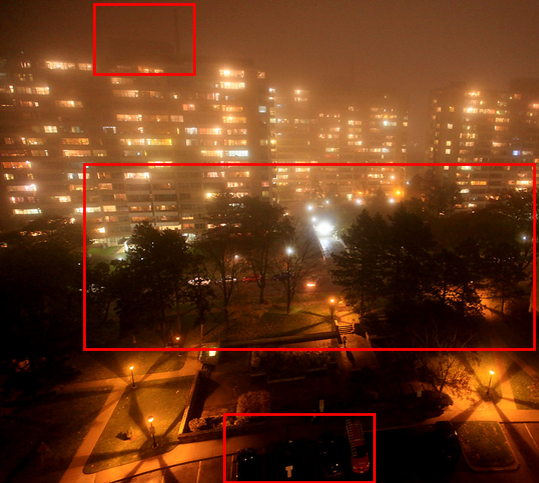}}\hfill
	\subfloat[Ours]{\includegraphics[width=0.245\textwidth]{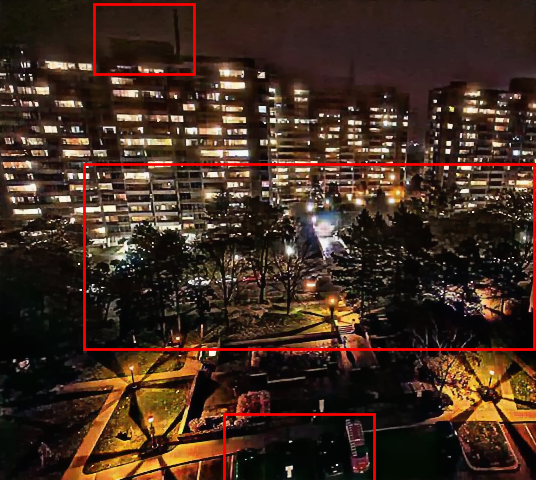}}\hfill
	\subfloat[Liu-22~\cite{liu2022nighttime}]{\includegraphics[width=0.245\textwidth]{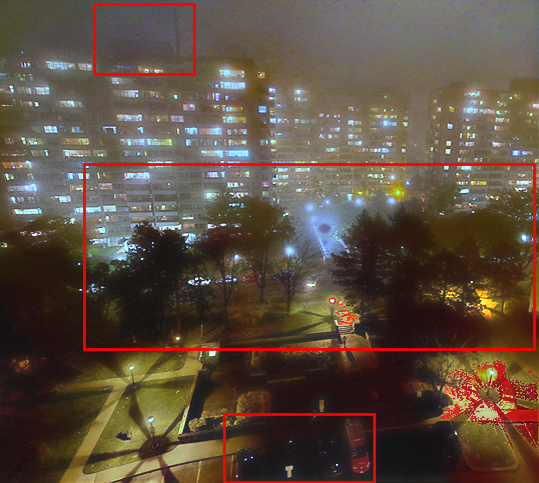}}\hfill
	\subfloat[Wang-22~\cite{wang2022variational}]{\includegraphics[width=0.245\textwidth]{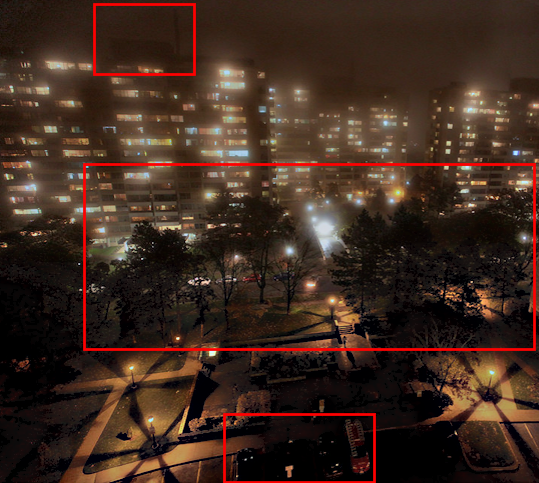}}\hfill
	\caption{Our nighttime dehazing results compared to existing methods, we can handle night glow and low-light conditions.}
	\label{fig:intro}
\end{teaserfigure}

\maketitle

\section{Introduction}
\label{sec:intro}
Nighttime hazy or foggy images often suffer from reduced visibility. In addition to the common issues faced by night images, such as low light, noise, uneven light distribution, and multiple light colors, nighttime hazy or foggy images also exhibit a strong glow and particle veiling effect.
Despite these challenges, addressing them is crucial for many applications. These include self-driving cars, autonomous drones, and surveillance~\cite{banerjee2021nighttime}, as haze during the nighttime are natural phenomena that are frequent and inevitable.

Daytime haze removal methods cannot handle the unique challenges posed by nighttime haze.
Traditional non-learning daytime dehazing methods (\eg,~\cite{tan2008visibility,fattal2008single,he2011single,berman2016non}) rely on the haze imaging model~\cite{koschmieder1924theorie}. 
However, this model is not valid at night due to the presence of artificial light sources and the complexity of illumination colors.
As a result, unlike in daytime, we cannot assume a uniform atmospheric light color.
Moreover, this daytime haze model does not account for the visual appearance of glow.

Existing nighttime dehazing methods produce unsatisfactory dark visuals or unmitigated glow effects.
Non-deep learning methods (e.g.,~\cite{li2015nighttime,yang2018superpixel,wang2022variational,liu2023multi}) introduce certain constraints on glow. 
However, they struggle with dark results due to the imprecise decomposition of glow and background layers or the use of dark channel prior~\cite{he2010single} for dehazing.
The main challenge faced by learning-based methods is the absence of real-world paired training data, as obtaining clear ground truth images of hazy nighttime scenes that include glow and multiple light sources is intractable.
A learning-based method~\cite{zhang2020nighttime} has attempted to address this issue by utilizing synthetic data. However, this method is unable to effectively suppress glow since the synthetic dataset does not account for the glow effect.
A semi-supervised deep learning-based network~\cite{yan2020nighttime} suffers from artifacts and loss of low-frequency scene details.

In this paper, our goal is to enhance visibility in a single nighttime haze image by suppressing glow and enhancing low-light regions.
Our glow suppression includes two main parts: APSF-guided glow rendering and gradient adaptive convolution.
Our glow rendering method uses an APSF-guided approach to create glow effects for various light sources. We employ a light source aware network to detect the locations of light sources in images and then apply APSF-guided glow rendering to these sources. 
Our framework learns from the rendered images and thus can suppress glow effects in different light sources.
Our gradient adaptive convolution captures edges and textures from hazy images. To be specific, edges are obtained by computing the pixel differences~\cite{su2021pixel} between neighboring pixels, while the bilateral kernel~\cite{tomasi1998bilateral} is used to extract textures of images. Both edges and textures are then fed into our framework to enhance the image details. 
To enhance the visibility of non-light regions, we introduce a novel attention-guided enhancement module.
The hazy regions have low weights, while the dark regions have high weights in the attention map.
As shown in~\Cref{fig:intro}, our method not only handles glow effects but also enhances the low-light regions.

Overall, our contributions can be summarized as follows: 
\begin{itemize}
	\item To our knowledge, our method is the first learning-based network that handles night glow and low-light conditions in one go. 
    \item  We present a light source aware network and APSF-guided glow rendering to simulate glow effects from different light sources. By learning from the APSF-guided glow rendering data, our framework effectively suppresses glow effects in real-world hazy images.
	\item Since night images contain less contrast, we employ gradient-adaptive convolution for edge enhancement and the bilateral kernel for texture enhancement.
\end{itemize}
Extensive experiments on nighttime images demonstrate the effectiveness of our approach in quantitative and qualitative evaluations. 
Our method achieves 30.38dB of PSNR, which outperforms existing nighttime dehazing methods by 13$\%$.

\begin{figure*}[t!]
	\centering
	\captionsetup[subfigure]{labelformat=empty}
	{\includegraphics[width=0.99\textwidth]{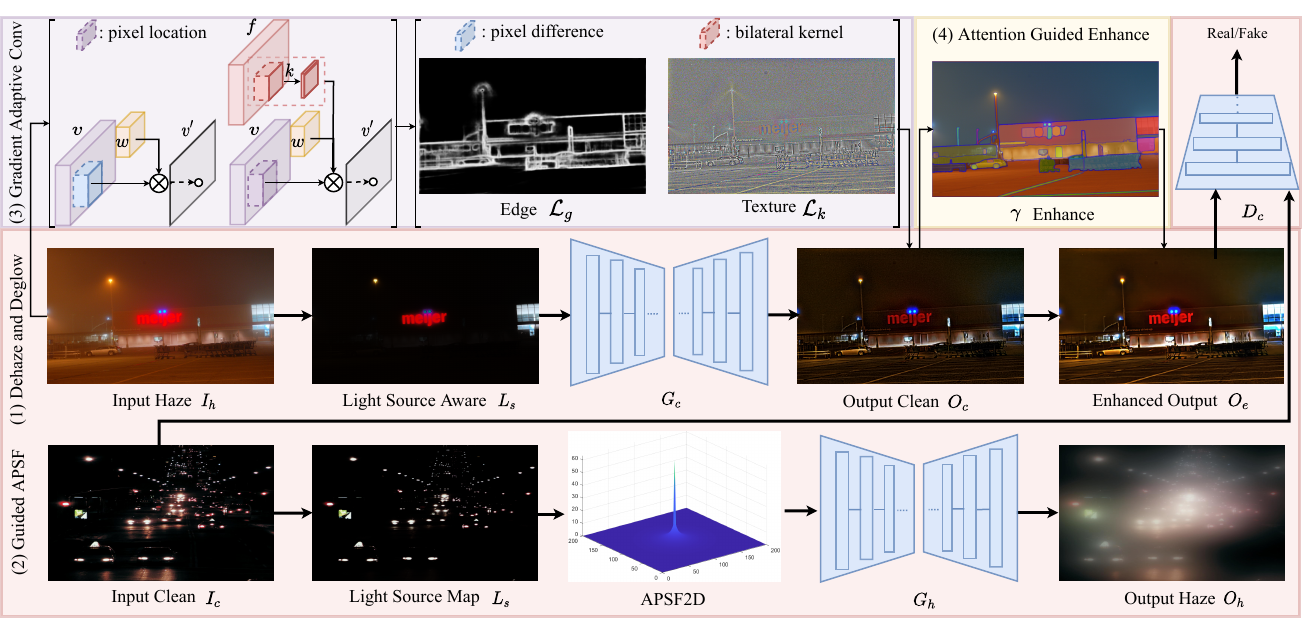}}
	\caption{(1) Our deglowing framework $G_c$ have two inputs: one to learn from real haze images $I_h$ and the other to learn from real clean reference images $I_c$. For input haze images $I_h$, $G_c$ output clean images $O_c$. For input clean images $I_c$, $G_c$ output clean images $G_c(I_c)$. 
		(2) APSF guide glow generator $G_h$ to generate glow $O_h$ on reference images $I_c$.
		(3) \textcolor{p}{the upper left} is the gradient adaptive convolution, from the gradient convolution \textcolor{b}{(the blue window)}, we obtain edges; from the adaptive bilateral kernel \textcolor{r}{(the red)}, we enhance texture details.
		(4) \textcolor{y}{the upper right} is attention-guided enhancement module.}
	\label{fig:framework}
\end{figure*}

\section{Related Work}
\label{sec:related_work}
Early dehazing methods utilized multiple images~\cite{narasimhan2000chromatic,li2015simultaneous} or priors for atmospheric light and transmission estimation~\cite{tan2008visibility, fattal2008single, he2011single,berman2018single}. 
With the advent of deep learning, numerous networks were proposed to estimate the transmission map~\cite{cai2016dehazenet, ren2016single, zhang2018densely} or output clean images end-to-end~\cite{li2017aod, ren2018gated, li2018single, qu2019enhanced,ye2022perceiving,wu2023ridcp,guo2023underwater,yu2022frequency,jin2022structure,ye2022towards}.
Recent fully supervised~\cite{liu2019griddehazenet, dong2020multi, qin2020ffa, zheng2021ultra, xiao2022single, wu2021contrastive, guo2022image,song2023vision,Wang_2023_ICCV}, semi-supervised~\cite{li2019semi, shao2020domain, chen2021psd, li2022physically,lin2024nightrain}, zero-shot~\cite{li2020zero, li2021you}, and unsupervised~\cite{huang2019towards, golts2019unsupervised, liu2020end, zhao2021refinednet, yang2022self} methods have been developed. 
However, these methods struggle with nighttime haze due to non-uniform, multi-colored artificial light and the absence of clean ground truth data for training.

Optimization-based nighttime dehazing methods have followed the atmospheric scattering model (\eg,~\cite{pei2012nighttime,ancuti2016night,ancuti2020day}), new imaging model (\eg,~\cite{zhang2014nighttime,zhang2017fast,tang2021nighttime,liu2022nighttime}), etc.
Pei and Lee~\cite{pei2012nighttime} transfer the airlight colors of hazy nighttime images to daytime and use DCP to dehaze.
Ancuti~\etal~\cite{ancuti2016night,ancuti2020day} introduce a fusion-based method and Laplacian pyramid decomposition to estimate local airlight.
Zhang~\etal~\cite{zhang2014nighttime} use illumination compensation, color correction and DCP to dehaze.
Zhang~\etal~\cite{zhang2017fast} propose maximum reflectance prior (MRP).
Tang~\etal~\cite{tang2021nighttime}	use Retinex theory and Taylor series expansion.
Liu~\etal~\cite{liu2023multi,liu2022single} use regularization constraints.
Wang~\etal~\cite{wang2022variational} proposed the gray haze-line prior and variational model.
Existing nighttime dehazing methods depend on local patch-based atmospheric light estimation, assuming uniformity within a small patch.
Therefore, their performance is sensitive to the patch size.
These methods are not adaptive and time-consuming in optimization.
Unlike them, our method is learning-based, more efficient, practical and fast.

Recently, learning-based nighttime dehazing methods~\cite{liu2023nighthazeformer} have been proposed.
Zhang~\etal~\cite{zhang2020nighttime} train the network using synthetic nighttime hazy images through fully supervised learning.
However, this approach does not account for glow, leaving it in the results.
Yan~\etal~\cite{yan2020nighttime} propose a semi-supervised method employing high-low frequency decomposition and a grayscale network.
However, their results tend to be dark, with lost details.
This is because coarse frequency-based decomposition methods struggle to effectively separate glow, leading to reduced brightness and visibility of the scene.
The DeGlow-DeHaze network~\cite{kuanar2022multi} estimates transmission followed by DehazeNet~\cite{cai2016dehazenet}.
However, the atmospheric light estimated by the DeHaze network is obtained from the brightest region and assumed to be globally uniform, which is invalid at nighttime~\cite{jin2023lighting,dai2022flare7k,dai2023flare7k++,Dai_2023_CVPR,dai2023mipi,sharma2020single,sharma2020nighttime,sharma2021nighttime}.
In contrast, our results can suppress glow and, at the same time, enhance low-light regions.

The glow of a point source, referred to as the Atmospheric Point Spread Function (APSF), has been studied in various works.
Narasimhan and Nayar~\cite{narasimhan2003shedding} first introduced APSF and developed a physics-based model for the multiple scattered light.
Metari~\etal~\cite{metari2007new} model the APSF kernel for multiple light scattering.
Li~\etal~\cite{li2015nighttime} decompose glow from the input image using a layer separation method~\cite{li2014single}, constrain glow by its smooth attribute, and dehaze using DCP.
Park~\etal~\cite{park2016nighttime} and Yang~\etal~\cite{yang2018superpixel} follow the nighttime haze model and use weighted entropy and super-pixel to estimate atmospheric light and transmission map.
However, these methods, after glow removal, simply apply daytime dehazing to nighttime dehazing, which results in low visibility and color distortion in their outcomes.
Previous works have primarily focused on optimization-based approaches, while our work is the first to incorporate the APSF prior into a nighttime learning network.

\section{Proposed Method}
\label{sec:method}
\Cref{fig:framework} shows our pipeline, including glow suppression and low-light enhancement. Our glow suppression has two parts: deglowing network $G_c$ and glow generator $G_h$.
Our deglowing network $G_c$ transforms real haze images $I_h$ to clean images $O_c$. We employ a discriminator $D_c$ to determine whether the generated clean images $O_c$ and the reference image $I_c$ are real or not.
Our novelty in the pipeline lies in these 3 ideas: APSF-guided glow rendering, gradient-adaptive convolution, and attention-guided enhancement.

\subsection{Light Source Aware Network}
\begin{figure*}[t!]
	\centering
	\captionsetup[subfigure]{labelformat=empty}
	{\includegraphics[width=0.99\textwidth]{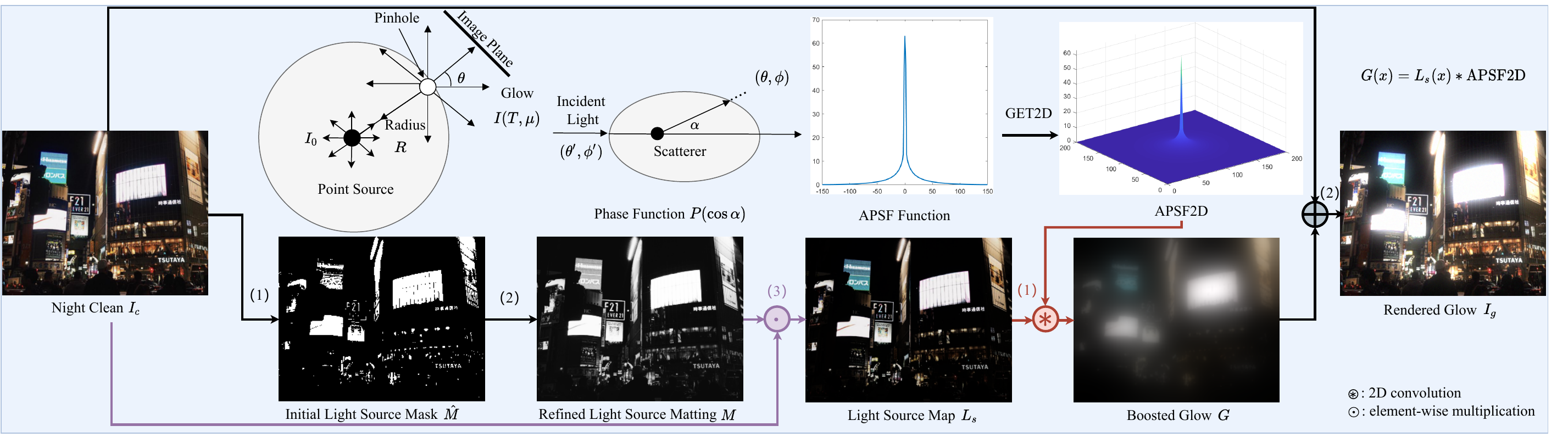}}
	\caption{We show~\cref{algorithm_ls}, light source map detection: (1) We first generate an initial light source mask $\hat{M}$ based on intensity, (2) then refine the mask using alpha matting~\cite{levin2007closed} to obtain light source soft matting ${M}$. \textcolor{p}{(3)} By multiplying the light source map ${M}$ with the night clean image $I_c$, we obtain the light source map $L_s$.
	After obtaining the light source, we show~\cref{algorithm_apsf}, APSF-guided nighttime glow rendering:
	 \textcolor{r}{(1)} Next, we perform APSF 2D convolution on the light source map to render glow $G$. (2) Finally, by combining the night clean and glow image, we obtain the rendered glow image $I_g$. More results are shown in~\Cref{fig:apsf}.}
	\label{fig:apsf_g}
\end{figure*}

\begin{algorithm}[t]
	\caption{Light Source Map Detection}
	\label{algorithm_ls}
	\begin{algorithmic}[1]
		\State \textbf{Generate an initial light source mask by thresholding the night input $I$,}
		$\hat{M}_{i,j} \gets \begin{cases} 1 & \text{if } \max_{c \in\{r, g, b\}}\left(I^c_{i,j}\right) > 0.8 \\ 0 & \text{otherwise} \end{cases}$
		\State \textbf{Refine $\hat{M}_{i,j}$ to ${M}_{i,j}$ using alpha matting~\cite{levin2007closed},}
		\State \textbf{Calculate the percentage of pixels in the mask,}
		\Statex $light\_sz \gets \frac{\sum(M)}{\text{numel}(M)} \times 100$
		\State \textbf{Obtain the light source image,} $L_s \gets I \odot M$
	\end{algorithmic}
\end{algorithm}

\begin{figure}[t]
	\centering
	\captionsetup[subfigure]{font=small, labelformat=empty}
	\captionsetup[subfloat]{farskip=1pt}
	\subfloat{\includegraphics[width = 0.245\columnwidth,height=1.5cm]{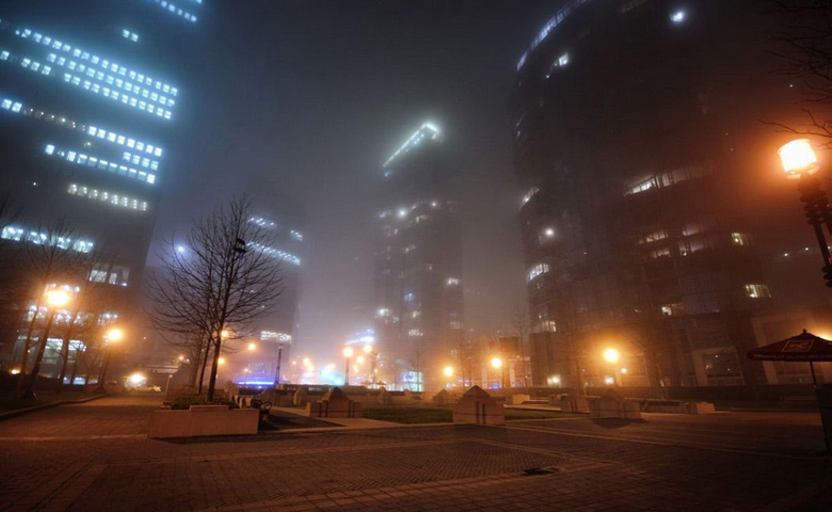}}\hfill
	\subfloat{\includegraphics[width = 0.245\columnwidth,height=1.5cm]{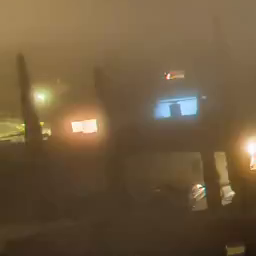}}\hfill
	\subfloat{\includegraphics[width = 0.245\columnwidth,height=1.5cm]{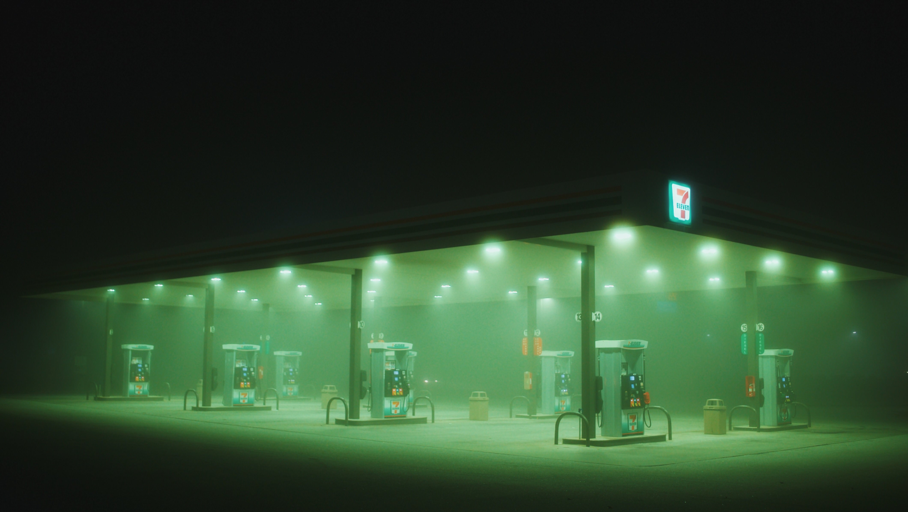}}\hfill
	\subfloat{\includegraphics[width = 0.245\columnwidth,height=1.5cm]{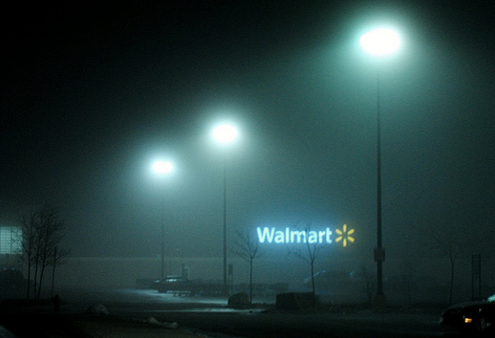}}\hfill\\	
	\subfloat{\includegraphics[width = 0.245\columnwidth,height=1.5cm]{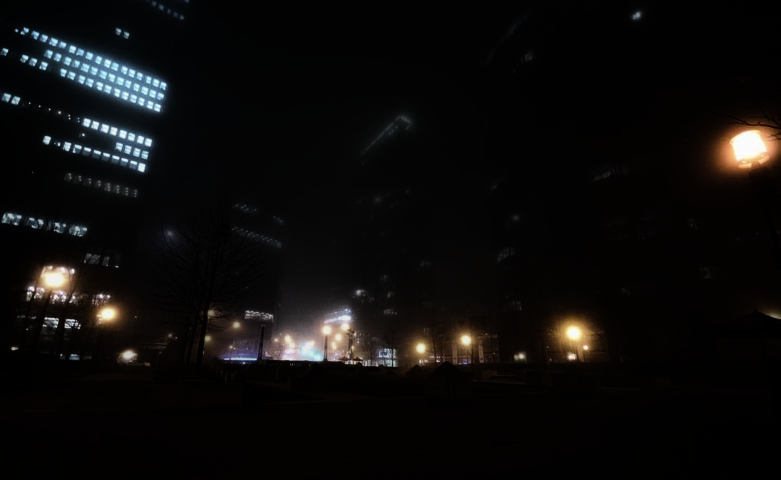}}\hfill
	\subfloat{\includegraphics[width = 0.245\columnwidth,height=1.5cm]{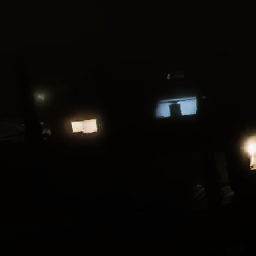}}\hfill
	\subfloat{\includegraphics[width = 0.245\columnwidth,height=1.5cm]{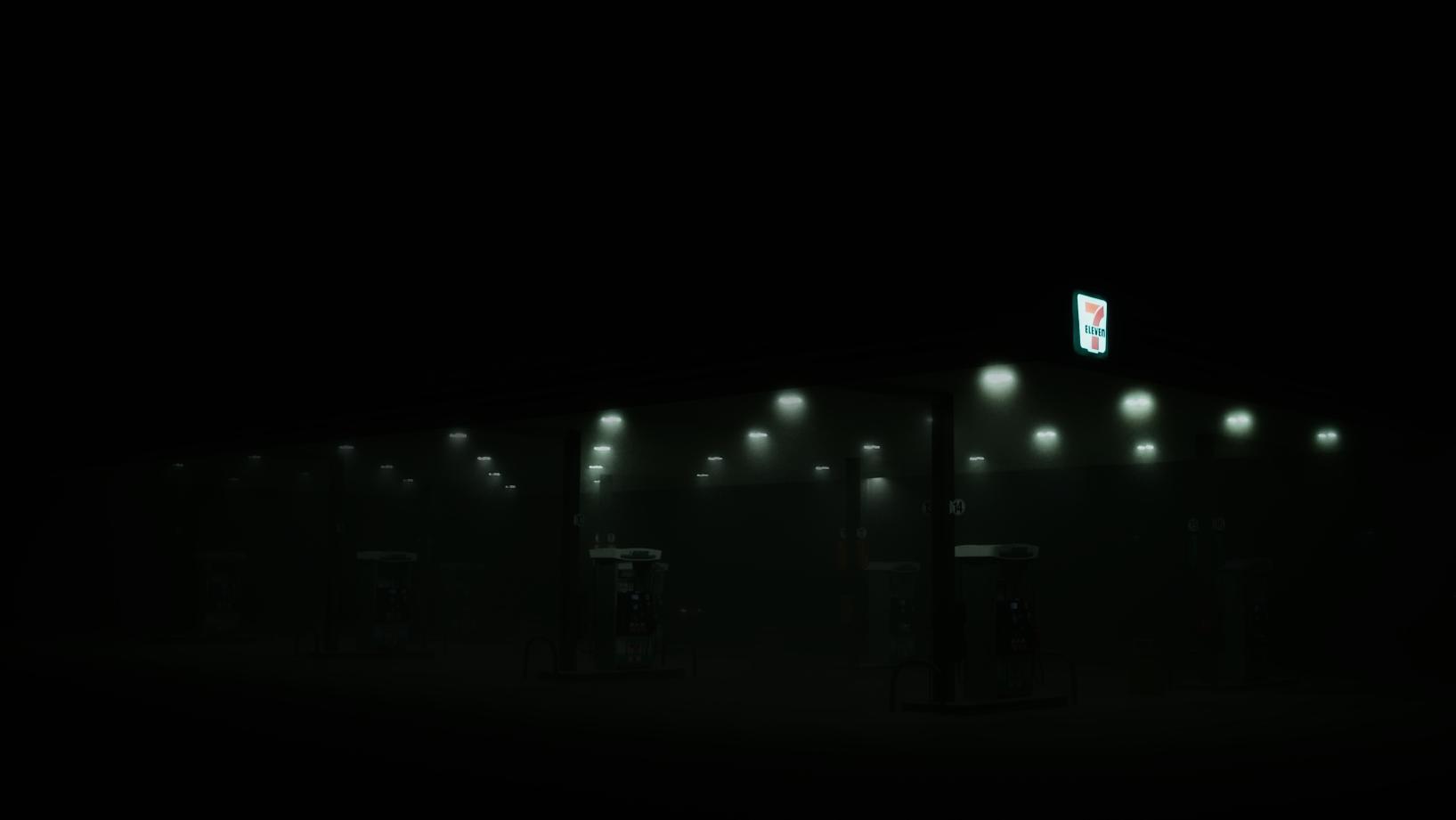}}\hfill
	\subfloat{\includegraphics[width = 0.245\columnwidth,height=1.5cm]{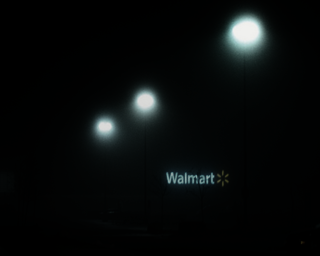}}\hfill
	\vspace{0.1cm}
	\caption{We show the light source maps $L_s$ of night haze.}
	\label{fig:ls_haze}
\end{figure}

\begin{figure}[t]
	\centering
	\captionsetup[subfloat]{farskip=1pt}
	\setcounter{subfigure}{0}
	\subfloat{\includegraphics[width=0.245\columnwidth,height=0.2\columnwidth]{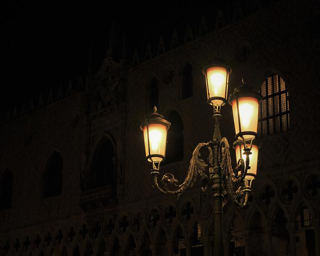}}\hfill
	\subfloat{\includegraphics[width=0.245\columnwidth,height=0.2\columnwidth]{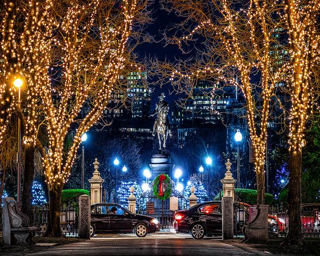}}\hfill
	\subfloat{\includegraphics[width=0.245\columnwidth,height=0.2\columnwidth]{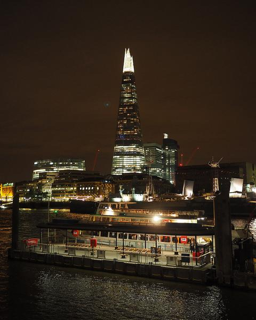}}\hfill
	\subfloat{\includegraphics[width=0.245\columnwidth,height=0.2\columnwidth]{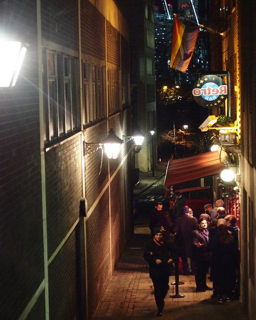}}\hfill
	\setcounter{subfigure}{0}			
	\subfloat{\includegraphics[width=0.245\columnwidth,height=0.2\columnwidth]{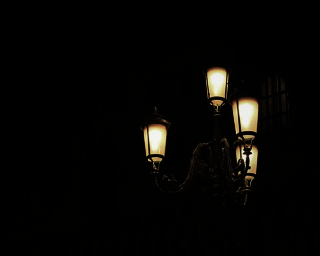}}\hfill
	\subfloat{\includegraphics[width=0.245\columnwidth,height=0.2\columnwidth]{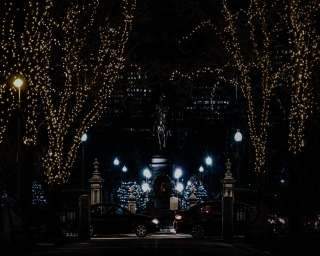}}\hfill
	\subfloat{\includegraphics[width=0.245\columnwidth,height=0.2\columnwidth]{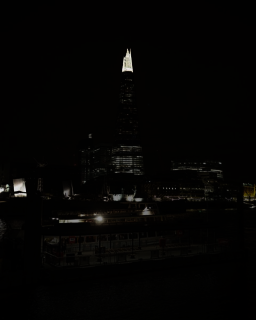}}\hfill
	\subfloat{\includegraphics[width=0.245\columnwidth,height=0.2\columnwidth]{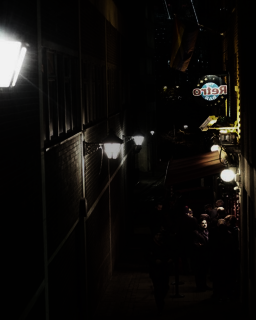}}\hfill	
	\setcounter{subfigure}{0}		
	\subfloat{\includegraphics[width=0.245\columnwidth,height=0.2\columnwidth]{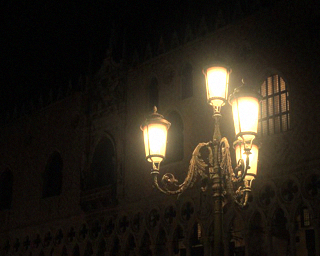}}\hfill
	\subfloat{\includegraphics[width=0.245\columnwidth,height=0.2\columnwidth]{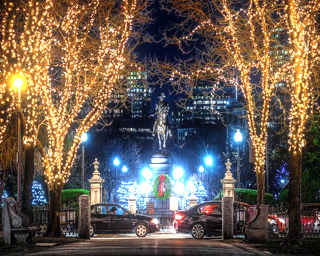}}\hfill
	\subfloat{\includegraphics[width=0.245\columnwidth,height=0.2\columnwidth]{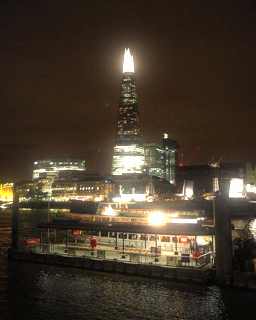}}\hfill
	\subfloat{\includegraphics[width=0.245\columnwidth,height=0.2\columnwidth]{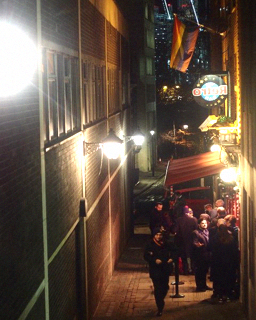}}\hfill
	\caption{We show with APSF, we can render glow $I_g$ (bottom) on night clean $I_c$ (top), with the help of light source maps.}
	\label{fig:apsf}
\end{figure}

Nighttime scenes often contain active light sources such as streetlights, car headlights, and building lights. These sources can cause strong glow in hazy nighttime scenes. The appearance of haze and glow in nighttime scenes can be modeled as ~\cite{li2015nighttime}:
\begin{equation}
\begin{aligned}
I_h(x) = I_c(x)t(x) + A(x)\left(1-t(x)\right) + L_s(x) * \text{APSF},
\label{eq:gmodel}
\end{aligned}
\end{equation}
where $G(x) = L_s(x) * \text{APSF}$, is the glow map, $L_s(x)$ represents the light sources, and APSF stands for Atmospheric Point Spread Function, $*$ denotes the 2D convolution.
$I_h$ is an observed hazy image. $I_c$ is the scene radiance (without haze). $A$ is the atmospheric light, and $t$ is the transmission, modeled as $t(x) = e^{-\beta d(x)}$, where $\beta$ is the extinction coefficient.
Light sources play an important role and can be utilized in three ways: (1) to inform $G_c$ about the location of light sources (as shown in~\Cref{fig:ls_haze}), (2) to guide $G_h$ to generate glow output $O_h$, and (3) to render the glow data $I_g$ using APSF (as shown in~\Cref{fig:apsf}).
For (1), we define light source consistency loss to keep the light source regions consistent in input and output images, therefore to maintain the same color and shape of these regions: 
\begin{align}
\mathcal{L}_\text{ls}= |O_c\odot M - L_s|_{1},
\label{eq:loss_1s}
\end{align}
where $L_s$ is the light source, $O_c$ is the output clean image, $M$ is the soft matting map, $\odot$ is element-wise multiplication.
The process of obtaining them is depicted in~\Cref{fig:apsf_g} and~\Cref{algorithm_ls}, which involves the following steps: First, we identify the regions that contain light sources. Next, an initial light source mask $\hat{M}$ is generated by thresholding the night image. To obtain a more accurate separation of the light sources from the surrounding areas and to ensure smoother transitions, we refine the initial mask $\hat{M}$ to a matting map $M$ using alpha matting~\cite{levin2007closed}. Finally, we isolate the light sources $L_s$ by applying the element-wise multiplication of the night image $I$ and the refined soft matting map $M$.

\begin{algorithm}[t]
	\caption{APSF-based Nighttime Glow Rendering}
	\label{algorithm_apsf}
	\begin{algorithmic}[1]
		\State \textbf{Compute the APSF function in~\Cref{algorithm_w},}
		\Statex $APSF \gets$ \Call{psfweight} {$\theta, T, q$}
		\State \textbf{Convert the APSF function to an APSF2D,}
		\Statex $APSF2D \gets$ \Call{get2D} {$APSF$}
		\State \textbf{Perform APSF2D convolution on the light source image,}
		\Statex  $G \gets L_s \ast APSF2D$,
		\State \textbf{Calculate the parameter for combining the clean and glow image,} $\epsilon \sim \mathcal{N}(0,1),$
		\Statex $\alpha \gets 0.4196 \cdot light\_sz^2 - 4.258 \cdot light\_sz + 11.35 + 0.05 \cdot \epsilon$,
		\State \textbf{Combine the clean and glow image to render night glow,}
		\Statex $I_g \gets 0.99 \cdot I_c + \alpha \cdot G$
		\State \textbf{Add Gaussian noise,}
		$I_g \gets$ \Call{AddNoise}{$I_g$}
	\end{algorithmic}
\end{algorithm}

\subsection{APSF-Guided Nighttime Glow Rendering}
After obtaining the light source maps, we present an APSF-based method for rendering nighttime glow effects, as shown in~\Cref{fig:apsf_g} and~\Cref{algorithm_apsf}. First, we compute the APSF function in~\cref{algorithm_w} and convert it into a 2D format, which allows us to perform 2D convolution on the light source maps, resulting in the glow images. Then, we combine the clean and glow images to render a realistic glow effect in the final image. In~\Cref{fig:apsf}, we provide examples of paired nighttime clean $I_c$ and glow $I_g$ images, demonstrating the effectiveness of our approach.

In~\Cref{fig:apsf_g} (top left), we use the phase function $P(\cos\theta)$~\cite{ishimaru1978wave} to approximate light scattering in nighttime glow. The scattering angle $\theta$ is defined as the angle between the incident $(\theta', \mu')$ and scattered light $(\theta, \mu)$ directions, where $\mu'=\cos\theta'$ and $\mu=\cos\theta$ represent the angular distribution of light. The scattered light intensity $I (T, \mu)$, considering optical thickness ($T$), measures the attenuation of light due to fog. Here, $T = \beta R$, with $\beta$ representing the attenuation coefficient and $R$ representing the distance between the isotropic source and the pinhole camera. The shape of the phase function $P$ depends on the size of the scatterer, and hence the weather conditions. For large particles, such as fog, $P$ exhibits a strong peak in the direction of light incidence.

\Cref{algorithm_w} shows the APSF weight calculation process~\cite{narasimhan2003shedding}.
There are three input parameters: angle $\theta$ (-180$^{\circ}$ to 180$^{\circ}$), optical thickness $T$, and forward scattering parameter $q$ that indicates the  atmospheric conditions~\cite{middleton1957vision}.
The phase function $P$ can be expanded using Legendre polynomials series $L_m$ ($m$ stand for order)~\cite{chandrasekhar2013radiative}:
\begin{equation}
\begin{aligned}
I(T, \mu)  =\sum_{m=1}^{\infty} g_m(T)\left(L_{m-1}(\mu)+L_m(\mu)\right),\label{Eq:phase}
\end{aligned}
\end{equation}
where: 
$g_m(T) =I_0 e^{-\beta_m T-\alpha_m \log T},$
$\alpha_m = m+1, \beta_m=\frac{2 m+1}{m}\left(1-q^{m-1}\right)$.

\begin{algorithm}[t]
	\caption{APSF Weights Calculation}
	\label{algorithm_w}
	\begin{algorithmic}[1]
		\Require $\theta$, optical thickness $T$, forward scattering parameter $q$ 
		\Ensure weights
		\State $\mu \gets \cos(\theta)$
		\If{look-up table (LUT) for Legendre polynomials exists}
		\State Load LUT
		\Else
		\State Generate LUT and save to file
		\EndIf
		\State Define $\alpha_\text{m} (m) \gets m + 1$, $\beta_\text{m} (m, q) \gets \frac{(2m + 1)(1 - q^{m-1})}{m}$
		\State Define $g_\text{m} (m, \alpha_\text{m}, \beta_\text{m}, T) \gets e^{(-\beta_\text{m} T - \alpha_\text{m}  \log(T))}$
		\State Initialize $weights$ to all zeros
		\For{$m \gets 1$ to 200 (number of polynomials)}
		\If{$m == 1$}
		\State $L_\text{m-1}(\mu) \gets 1$
		\Else
		\State $L_\text{m-1}(\mu) \gets L_\text{m}(\mu)$
		\EndIf
		\State $L_\text{m}(\mu) \gets$ interpolate LUT for $m$ and $\mu$
		\State $p \gets g_\text{m}(m, \alpha_\text{m}(m), \beta_\text{m}(m, q), T)(L_\text{m-1}(\mu) + L_\text{m}(\mu))$,~\cref{Eq:phase}.
		\State $weights \gets weights + p$
		\EndFor
		\State $weights \gets weights \cdot T^2$ (normalize) 
	\end{algorithmic}
\end{algorithm}

\subsection{Gradient Adaptive Convolution}
\begin{figure}[t]
	\centering
	\captionsetup[subfloat]{farskip=1pt}
	\setcounter{subfigure}{0}
	\subfloat{\includegraphics[width=0.245\columnwidth,height=0.2\columnwidth]{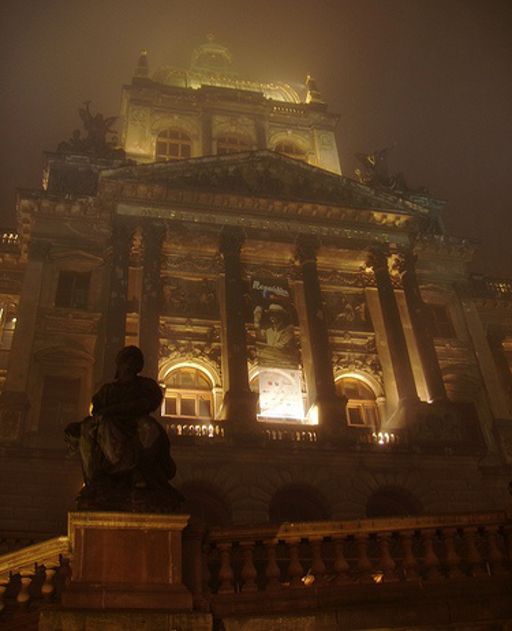}}\hfill
	\subfloat{\includegraphics[width=0.245\columnwidth,height=0.2\columnwidth]{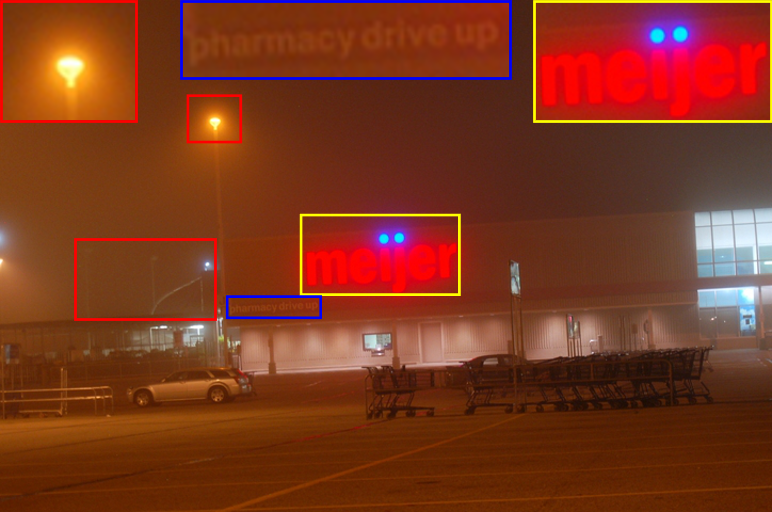}}\hfill
	\subfloat{\includegraphics[width=0.245\columnwidth,height=0.2\columnwidth]{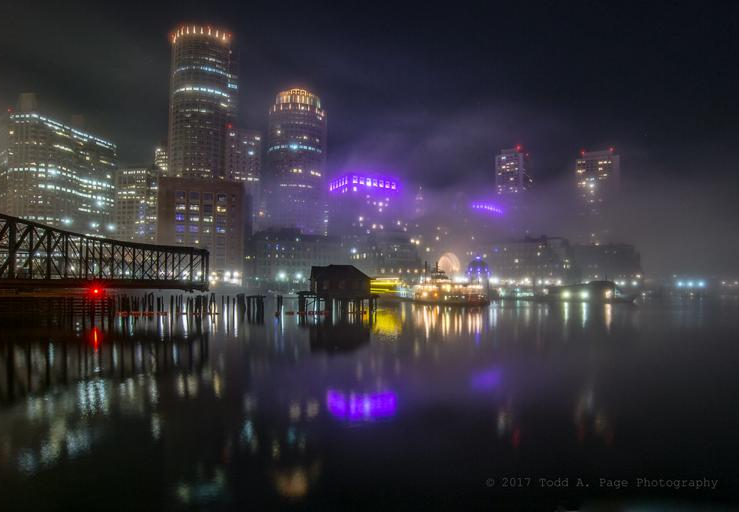}}\hfill
	\subfloat{\includegraphics[width=0.245\columnwidth,height=0.2\columnwidth]{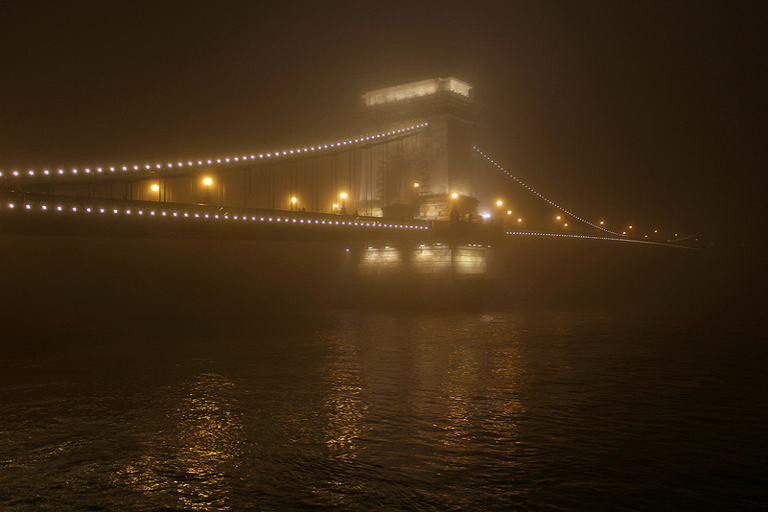}}\hfill
	\setcounter{subfigure}{0}
	\subfloat{\includegraphics[width=0.247\columnwidth,height=0.2\columnwidth]{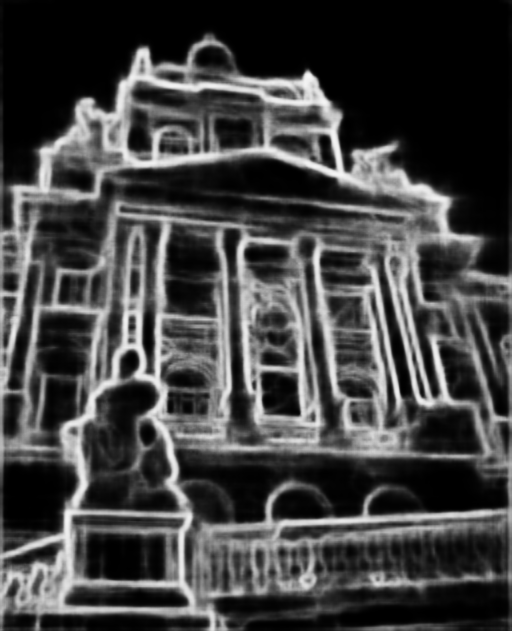}}\hfill
	\subfloat{\includegraphics[width=0.247\columnwidth,height=0.2\columnwidth]{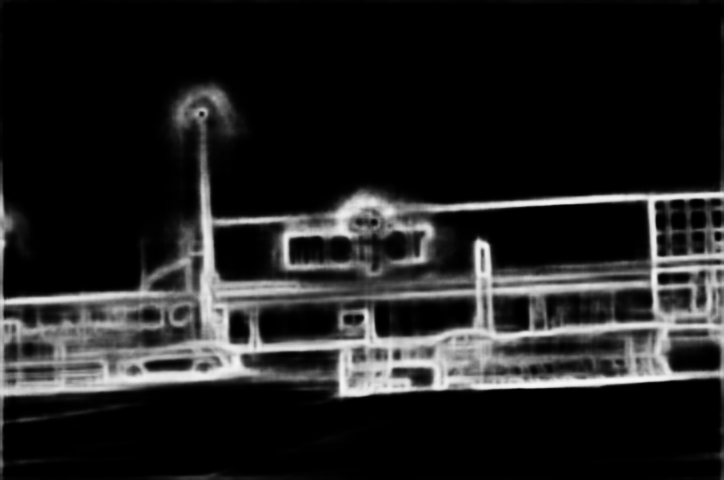}}\hfill
	\subfloat{\includegraphics[width=0.247\columnwidth,height=0.2\columnwidth]{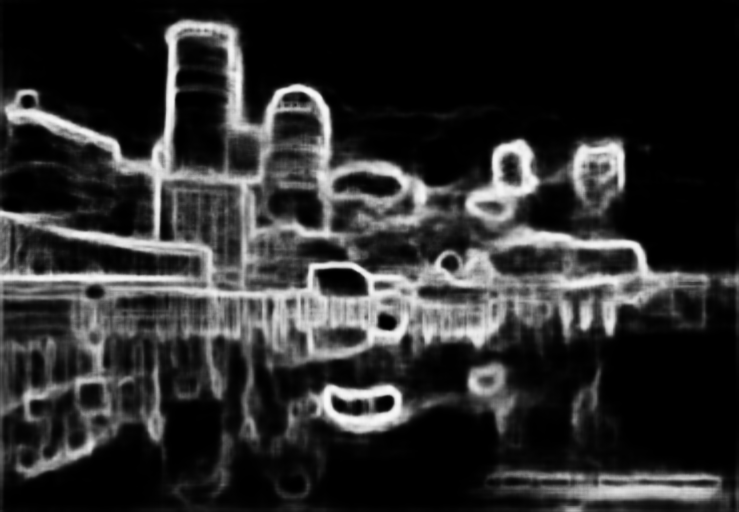}}\hfill
	\subfloat{\includegraphics[width=0.247\columnwidth,height=0.2\columnwidth]{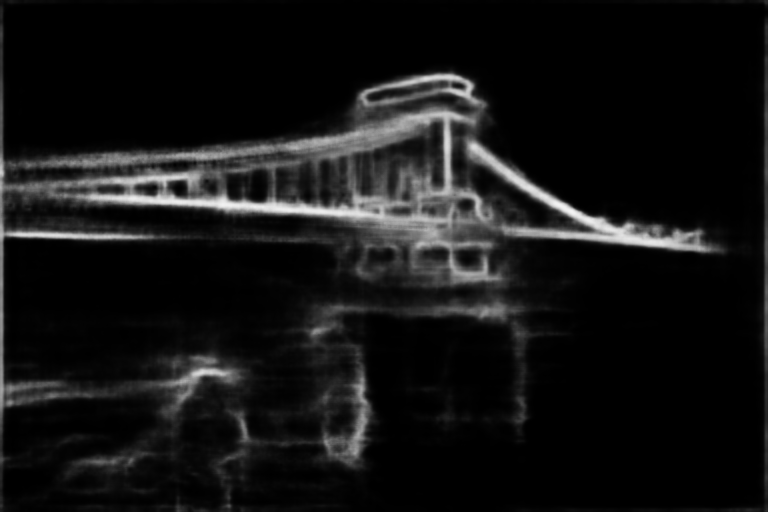}}\hfill
	\setcounter{subfigure}{0}
	\subfloat{\includegraphics[width=0.247\columnwidth,height=0.2\columnwidth]{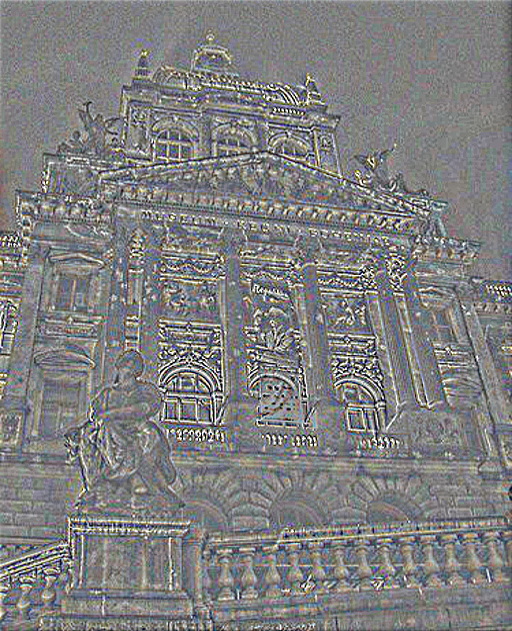}}\hfill
	\subfloat{\includegraphics[width=0.247\columnwidth,height=0.2\columnwidth]{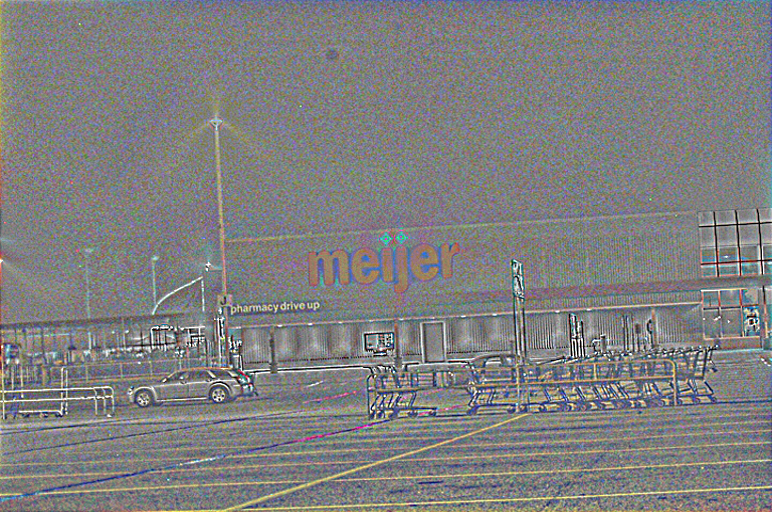}}\hfill
	\subfloat{\includegraphics[width=0.247\columnwidth,height=0.2\columnwidth]{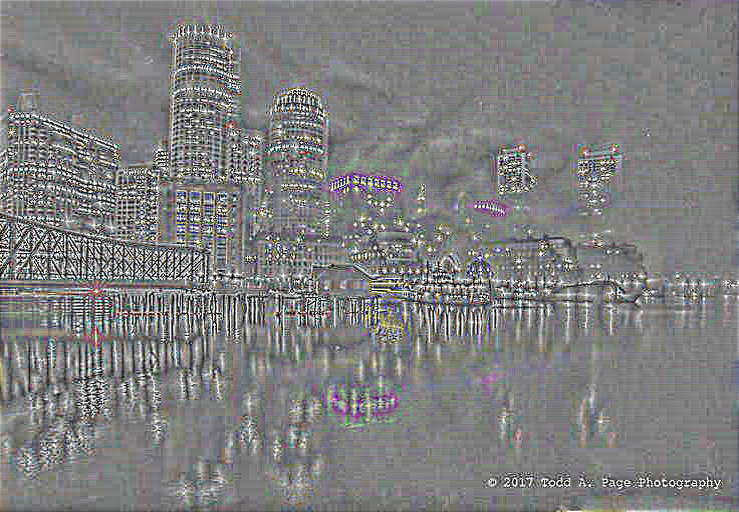}}\hfill
	\subfloat{\includegraphics[width=0.247\columnwidth,height=0.2\columnwidth]{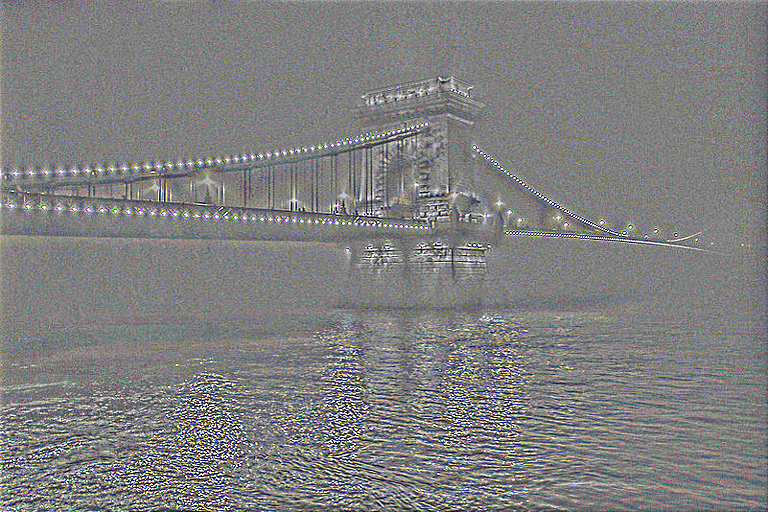}}\hfill
	\caption{We show using gradient adaptive convolution, can obtain edges (middle) and textures (below) in haze images.}
\label{fig:edge_texture}
\end{figure}

Nighttime hazy images suffer from low contrast, missing textures, and uneven illumination. Vanilla 2D convolution, which computes the output based solely on input pixel values and the weights in the convolution kernel, may not be sufficient for enhancing nighttime haze images. Hence, to improve the edge and texture details shown in~\Cref{fig:edge_texture}, we utilize gradient-adaptive convolution.

\begin{figure}[t!]
	\centering
	\captionsetup[subfigure]{labelformat=empty}
	{\includegraphics[width=0.48\textwidth]{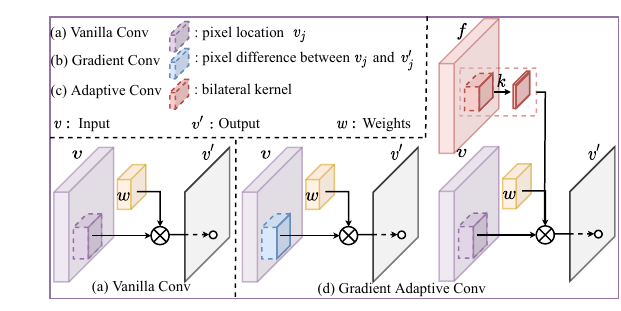}}
	\caption{Our gradient adaptive convolution combine (b) gradient convolution (\textcolor{b}{blue}) to capture edge, and (c) adaptive bilateral kernel (\textcolor{r}{red}) to obtain texture.}
	\label{fig:gac}
\end{figure}

\subsubsection{Edge Capture using Gradient Convolution}
We utilize nearby information to preserve the gradient information and improve contrast by extracting edges using gradient-adaptive convolution. The main idea of gradient convolution is to use local binary patterns (LBP)~\cite{pietikainen2010local,su2021pixel} in a $k\times k$ neighborhood, where the central pixel serves as a threshold and the values of the surrounding pixels are compared to the central pixel. If a neighboring pixel's value exceeds the central pixel's value, a position marker of 1 is assigned; otherwise, a value of 0 is assigned.
The vanilla convolution and gradient convolutions are expressed as follows:
\begin{align}
\mathbf{v}_i^{\prime}& =\sum_{j \in \Omega(i)} \mathrm{w}\left[\mathrm{p}_i-\mathrm{p}_j\right] \mathbf{v}_j,\;\;\;\;\;\;\;\;\;\;\;\;\;\;\;\;\;\;\;\;\ \text{(Vanilla Conv)} \label{conv}\\
\mathbf{v}_i^{\prime}& =\sum_{j \in \Omega(i)} \mathrm{w}\left[\mathrm{p}_i-\mathrm{p}_j\right] (\mathbf{v}_j- \mathbf{v}_j'), \;\;\;\;\,\;\;\;\ \text{(Gradient Conv)} \label{gconv}
\end{align}
where $\mathbf{v}=(\mathbf{v}_1, \dots, \mathbf{v}_n), \mathbf{v}_i \in \mathbb{R}^{c}$ over $n$ pixels and $c$ channels, is input image features.
$\mathrm{w}$ is the weight in the $\Omega(\cdot)$, $k \times k$ convolution window.
We use $[\mathbf{p}_i - \mathbf{p}_j]$ to denote indexing of the spatial dimensions of an array with 2D spatial offsets. 
This convolution operation results in a $c'$-channel output,
$\mathbf{v}'_i \in \mathbb{R}^{c'}$, at each pixel $i$.
In~\cref{fig:gac} (a) and (b), instead of using input pixel values in~\cref{conv}, we use the differences between neighboring pixels in~\cref{gconv} to capture the gradient information.

We fine-tune the pre-trained pixel difference network~\cite{su2021pixel} on nighttime haze images. Then, we use this fine-tuned network $D$ to extract edges from input haze images and enforce the consistency between the input $I_h$ and the output $O_c$. Our self-supervised gradient loss is defined as follows:
\begin{equation}
\mathcal{L}_g = |D(O_c) - D(I_h)|_{1}.
\label{eq:e_loss}
\end{equation} 

\subsubsection{Texture Capture using Bilateral Kernel}
A single CNN may not be sufficient for effectively handling uneven light distribution within a nighttime haze image. Therefore, we propose using a bilateral kernel in adaptive convolution to extract texture details, as shown in~\cref{fig:gac} (c):
\begin{align} 
\mathbf{v}_i^{\prime}& =\sum_{j \in \Omega(i)} K\left(\mathbf{f}_i, \mathbf{f}_j\right) 
\mathbf{w}\left[\mathbf{p}_i-\mathbf{p}_j\right] \mathbf{v}_j, \;\;\;\;\;\,\text{(Adaptive Conv)}\label{aconv}
\end{align}
where $K$ is bilateral kernal, depends on pixel features $\mathbf{f}$~\cite{su2019pixel}.
We use color features $\mathbf{f} =(r, g, b)$,
$K\left(\mathbf{f}_i, \mathbf{f}_j\right) =\exp \left(-\frac{1}{2 \alpha_1}\left\|\mathbf{f}_i-\mathbf{f}_j\right\|^2\right)$,
$\mathbf{w}\left[\mathrm{p}_i-\mathrm{p}_j\right] =\exp \left(-\frac{1}{2 \alpha_2}\left\|\mathrm{p}_i-\mathrm{p}_j\right\|^2\right)$.

Since the features obtained using the bilateral kernel are noise-reduced, edge-preserved, and detail-enhanced, they help extract high-frequency texture details that are less affected by haze and glow, as shown in ~\Cref{fig:edge_texture} (bottom). To enforce the consistency between the input $I_h$ and the output $O_c$ and improve the quality of the output, we define the self-supervised bilateral kernel loss as:
\begin{equation}
\mathcal{L}_k = |K(O_c) - K(I_h)|_{1}.
\label{eq:k_loss}
\end{equation}
In summary, our gradient-adaptive convolution captures edge information by considering the differences between neighboring pixels instead of their pixel values, as reflected in the gradient term $(\mathbf{v}_j - \mathbf{v}_j')$. Additionally, it accounts for spatially varying illumination and non-uniform haze distribution by utilizing the adaptive term $K(\mathbf{f}_i, \mathbf{f}_j)$, which helps to preserve texture information and adapt more effectively to nighttime scenes.

\subsection{Network and Other Losses}
In~\Cref{fig:framework}, our glow suppression is CycleGAN~\cite{CycleGAN2017}-based network. We use the deglowing network $G_c$, which is coupled with a discriminator $D_{c}$. 
To ensure reconstruction consistency, we have another haze generator $G_{h}$ coupled with its discriminator $D_{h}$. 
$G_c$ input hazy images $I_h$ and output clean images $O_c$.
Our haze generator $G_h$ transforms the output images $O_c$ to reconstructed clean images $\hat{I}_h$.
By imposing the cycle-consistency constraints, our deglowing network $G_c$ learns to remove the real-world glow. 
Thanks to the cycle-consistency constraints, we are allowed to use unpaired data to optimize our deglowing network, thus reducing the domain gap between the synthetic datasets and real-world glow images.

Besides the self-supervised light source consistency loss $\mathcal{L}_\text{ls}$, gradient loss $\mathcal{L}_g$ and bilateral kernel loss $\mathcal{L}_k$, with weights \{1, 0.5, 5\}, we followed~\cite{CycleGAN2017}, use other losses to train our network. 
They are adversarial loss $\mathcal{L}_{adv}$, cycle consistency loss $\mathcal{L}_{cyc}$, identity loss $\mathcal{L}_{iden}$, with weights \{1, 10, 10\}. 

\subsection{Low-light Region Enhancement}
Nighttime dehazing often leads to dark results due to low-light conditions~\cite{jin2022unsupervised,wang2022low,wang2023exposurediffusion,guo2022enhancing,jin2022shadowdiffusion,jin2023estimating,wu2023generation}. To address this, we incorporate a low-light enhancement module to improve the visibility of object regions. Our approach involves generating an attention map that highlights these regions, allowing our method to focus on enhancing their intensity. We then apply a low-light image enhancement technique~\cite{guo2016lime} to enhance the region with the assistance of the attention map, even in scenes with low light.

\begin{figure}[t]
	\captionsetup[subfloat]{farskip=1pt}
	\setcounter{subfigure}{0}
	\subfloat{\includegraphics[width=0.245\columnwidth,height=0.2\columnwidth]{figures/Input/flickr3_real_B.png}}\hfill
	\subfloat{\includegraphics[width=0.245\columnwidth,height=0.2\columnwidth]{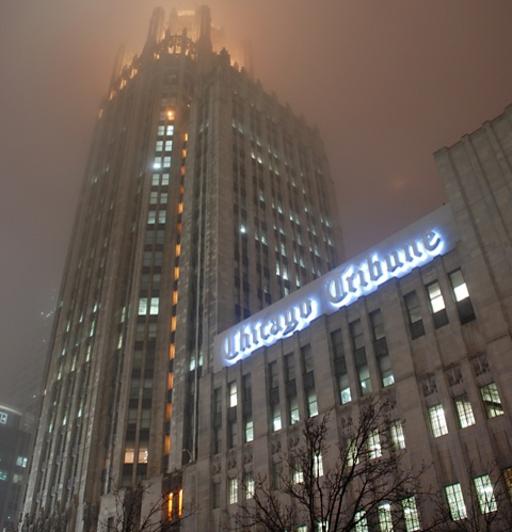}}\hfill
	\subfloat{\includegraphics[width=0.245\columnwidth,height=0.2\columnwidth]{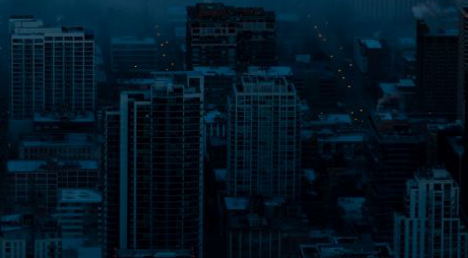}}\hfill
	\subfloat{\includegraphics[width=0.245\columnwidth,height=0.2\columnwidth]{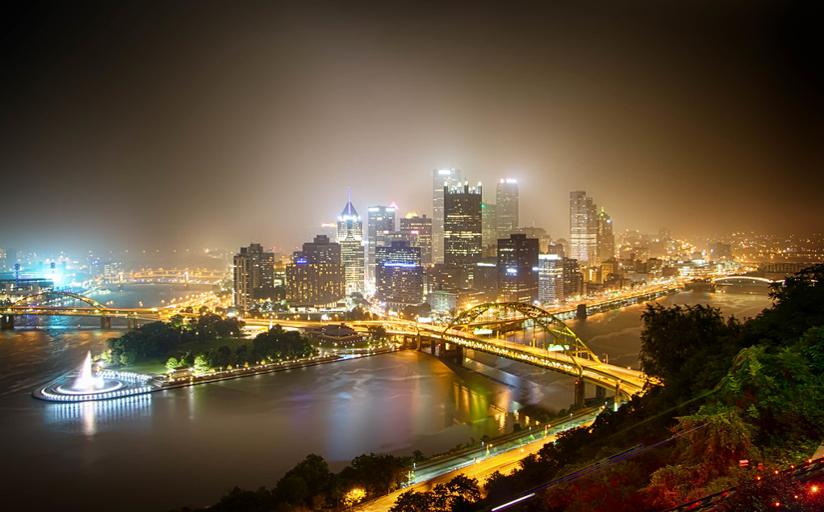}}\hfill\\
	\subfloat{\includegraphics[width=0.245\columnwidth,height=0.2\columnwidth]{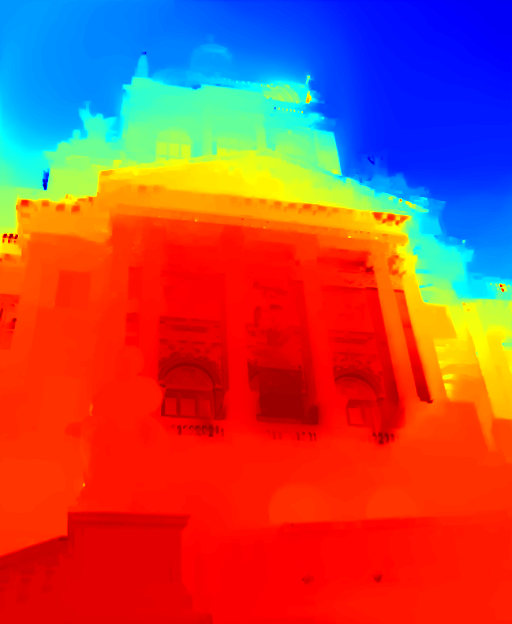}}\hfill
	\subfloat{\includegraphics[width=0.245\columnwidth,height=0.2\columnwidth]{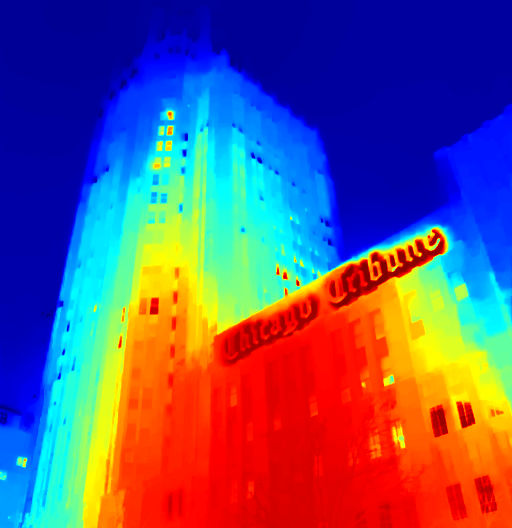}}\hfill
	\subfloat{\includegraphics[width=0.245\columnwidth,height=0.2\columnwidth]{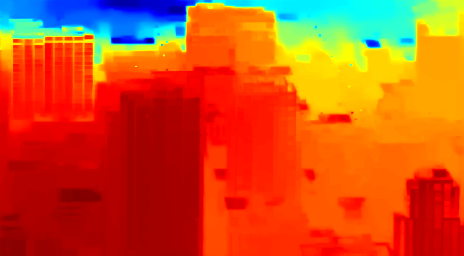}}\hfill
	\subfloat{\includegraphics[width=0.245\columnwidth,height=0.2\columnwidth]{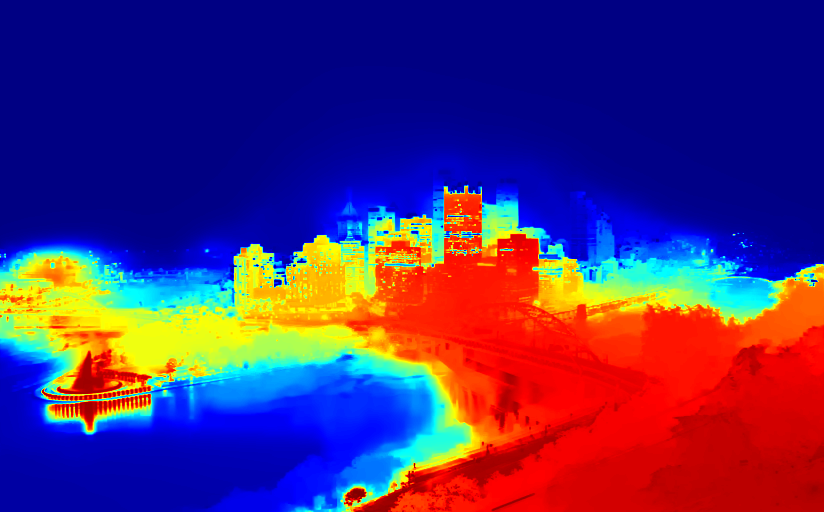}}\hfill
	\caption{We show attention maps to enhance low-light.}
	\label{fig:lowlight}
\end{figure}

\vspace{0.2cm}
\noindent \textbf{Attention Map}
To obtain the soft attention maps $A$ shown in~\Cref{fig:lowlight}, we input the night haze images and refine the coarse map using~\cite{li2014contrast}.
The refined attention map $A$ exhibits high values in object regions and low values in uniform regions, such as the sky.
Therefore, we can distinguish between the object and the haze regions.
\begin{equation}
O_e = (1 - A) \cdot O_c +  A \cdot O_c^\gamma,
\end{equation}
where $\gamma$ is the enhanced parameter in~\cite{guo2016lime}, we set 0.3 in our experiments, $O_c$ is the dehazed output, $O_e$ is the enhanced result, $A$ is the attention map.

\begin{table}[t!]
	\small
	\centering
	\caption {Results on GTA5 nighttime haze dataset.} \label{tb:tb_real_syn}
	\begin{tabularx}{\columnwidth}{ c|Y|Y }
		\hline
		\multirow{2}{*}{}   & PSNR $\uparrow$			  & SSIM $\uparrow$\\
		\hline                    
		Input Image            & 18.987 		  & 0.6764 \\ 
		\hline
		Li~\cite{li2015nighttime} 			   & 21.024 		  & 0.6394 \\
		\hline                    
		Zhang~\cite{zhang2017fast} 		   & 20.921 		  & 0.6461 \\
		\hline
		Ancuti~\cite{ancuti2016night} 		   & 20.585 		  & 0.6233 \\
		\hline	
		Yan~\cite{yan2020nighttime}    & 26.997 & 0.8499\\	\hline
		\hline
		CycleGAN~\cite{CycleGAN2017} 			   & 21.753 		  & 0.6960 \\
		\hline
		w/o  Gradient Adaptive Conv &  27.913      	  & 0.8673 \\
		\hline
		w/o APSF Guided Render& 28.913      	  &0.8776  \\
		\hline	
		\textbf{Our Result}    & \textbf{30.383} & \textbf{0.9042}\\
		\hline		
	\end{tabularx}  
\end{table}

\begin{table}[!t]
	\centering
	\renewcommand{\arraystretch}{1.2}
	\caption{Summary of comparisons between ours and existing learning-based nighttime dehazing methods. SL is short for supervised learning, SSL is semi-supervised learning.}
	\resizebox{0.99\columnwidth}{!}{
		\begin{tabular}{c|c|c|c|c|c|c}\hline
			Learning &Methods &\multicolumn{1}{c|}{Glow} & Edge, Texture & Low Light &Light Source &Uneven light\\\hline
			{\multirow{1}{*}{SSL}}  &Ours	   &\checkmark  &\checkmark &\checkmark &\checkmark &\checkmark\\\hline
			{\multirow{1}{*}{SL}}   &Kuanar~\cite{kuanar2022multi} &\checkmark  &$\times$ &$\times$ &$\times$ &$\times$\\\hline
			{\multirow{1}{*}{SSL}}  &Yan~\cite{yan2020nighttime}   & \checkmark  & \checkmark &$\times$ & $\times$  & $\times$ \\\hline
			{\multirow{1}{*}{SL}}   &Zhang~\cite{zhang2017fast}   & $\times$  & $\times$ & $\times$ & $\times$  & $\times$ \\\hline
	\end{tabular}}
	\label{tb:compare}
\end{table}

\begin{table*}[t]
	\footnotesize
	\centering
	\renewcommand{\arraystretch}{1.2}
	\caption{User study evaluation on the real night images, our method obtained the highest mean (the max score is 10), showing our method is effective in nighttime dehazing, deglowing and low-light enhancement. Our method is also visual realistic. The best result is in {\bf\textcolor{r}{red}} whereas the second and third best results are in {\textcolor{b}{blue}} and {\textcolor{p}{purple}}, respectively.}
	\resizebox{1\textwidth}{!}{
		\begin{tabular}{l|c|c|c|c|c|c|c|c|c|c}\hline 
			Aspects &Ours &Yan~\cite{yan2020nighttime} &Zhang~\cite{zhang2014nighttime} &Li~\cite{li2015nighttime} &Ancuti~\cite{ancuti2016night} &Zhang~\cite{zhang2017fast}  &Yu~\cite{yu2019nighttime} &Zhang~\cite{zhang2020nighttime} &Liu-22~\cite{liu2022nighttime} &Wang-22~\cite{wang2022variational} \\\hline 
			1.Dehaze$\uparrow$  &$\bf\textcolor{r}{9.1}\pm0.99$ &$\textcolor{b}{8.9}\pm1.75$ &$4.2\pm2.42$  &$\textcolor{p}{6.1}\pm2.00$ &$5.2\pm2.35$ &$4.4\pm2.01$ &$4.8\pm2.26$ &$4.6\pm2.07$ &$3.9\pm2.12$ &$5.5\pm1.91$\\\hline
			2.Deglow$\uparrow$  &$\bf\textcolor{r}{9.1}\pm0.91$ &$\textcolor{b}{7.9}\pm1.12$ &$3.2\pm1.96$  &$\textcolor{p}{6.2}\pm1.83$  &$5.2\pm2.14$ &$4.0\pm2.09$ &$3.7\pm2.08$ &$4.4\pm2.05$ &$5.3\pm1.99$ &$5.7\pm1.76$\\\hline
			3.Low-light$\uparrow$ &$\bf\textcolor{r}{8.5}\pm1.30$ &$\textcolor{b}{7.9}\pm1.93$ &$\textcolor{p}{7.1}\pm2.31$  &$5.5\pm2.33$  &$5.4\pm2.01$ &$5.6\pm1.82$ &$6.5\pm2.12$ &$5.6\pm1.75$ &$5.5\pm1.92$ &$5.4\pm1.89$\\\hline
			4.Realistic$\uparrow$   &$\bf\textcolor{r}{8.9}\pm0.94$ &$\textcolor{b}{8.0}\pm1.28$ &$4.6\pm1.93$  &$4.7\pm2.07$  &$\textcolor{p}{6.7}\pm1.90$ &$4.9\pm1.93$ &$5.7\pm1.90$ &$4.9\pm1.97$ &$3.8\pm1.81$ &$5.9\pm1.67$\\\hline
	\end{tabular}}
	\label{tb:tb_user}
\end{table*}

\begin{figure*}
	\centering
	\captionsetup[subfloat]{farskip=1pt}
	\setcounter{subfigure}{0}	
	\subfloat[Input]{\includegraphics[width=0.245\textwidth]{figures/Input/511.png}}\hfill
	\subfloat[Ours]{\includegraphics[width=0.245\textwidth]{figures/Ours/511.png}}\hfill
	\subfloat[Liu-22~\cite{liu2022nighttime}]{\includegraphics[width=0.245\textwidth]{figures/Liu/511.png}}\hfill
	\subfloat[Wang-22~\cite{wang2022variational}]{\includegraphics[width=0.245\textwidth]{figures/TIP22/511.png}}\hfill
	\subfloat[Zhang~\cite{zhang2020nighttime}]{\includegraphics[width=0.245\textwidth]{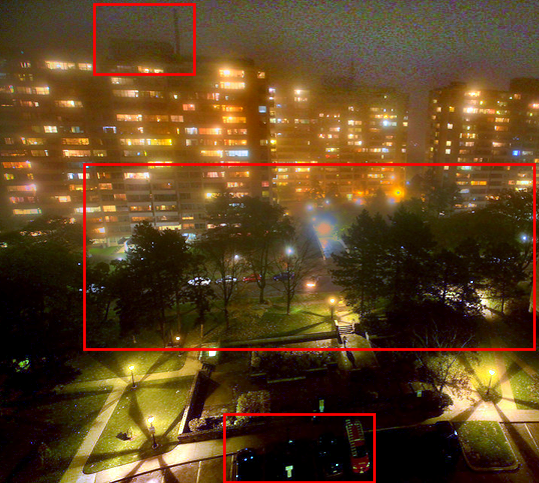}}\hfill
	\subfloat[Yan-20~\cite{yan2020nighttime}]{\includegraphics[width=0.245\textwidth]{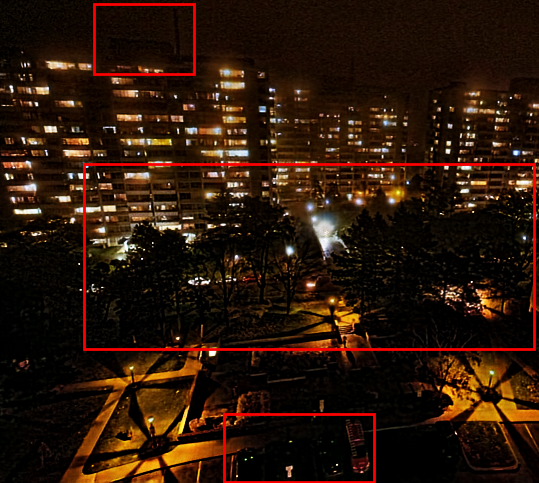}}\hfill
	\subfloat[Yu-19~\cite{yu2019nighttime}]{\includegraphics[width=0.245\textwidth]{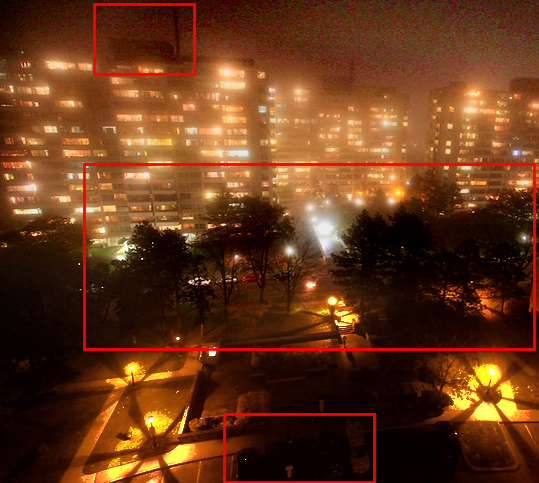}}\hfill
	\subfloat[Zhang~\cite{zhang2017fast}]{\includegraphics[width=0.245\textwidth]{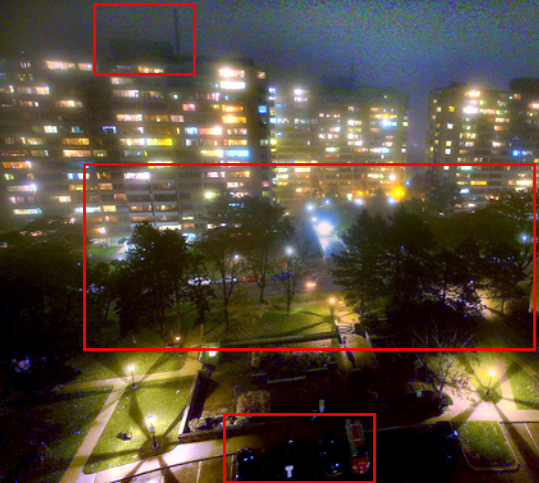}}\hfill
	\vspace{0.03in}
	\setcounter{subfigure}{0}
	\subfloat[Input]{\includegraphics[width=0.245\textwidth]{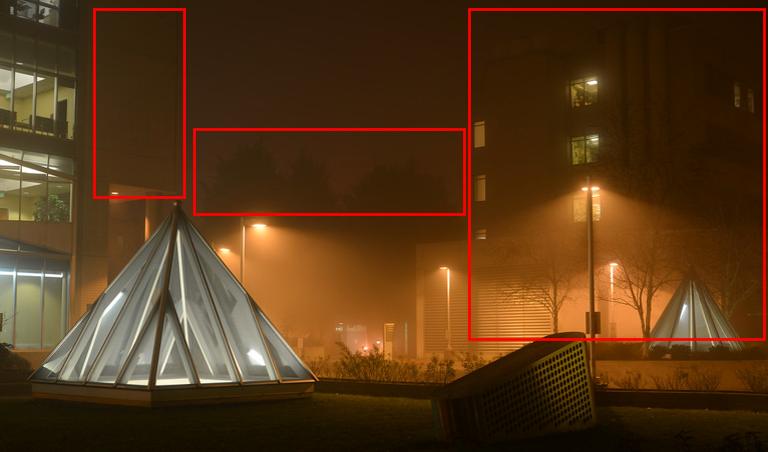}}\hfill
	\subfloat[Ours]{\includegraphics[width=0.245\textwidth]{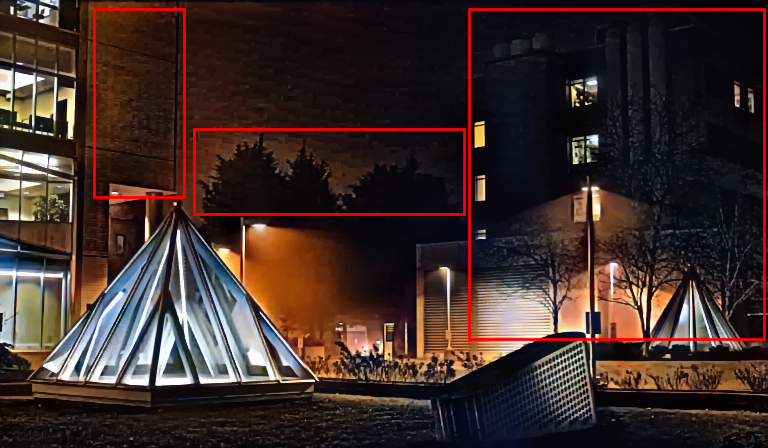}}\hfill
	\subfloat[Liu-22~\cite{liu2022nighttime}]{\includegraphics[width=0.245\textwidth]{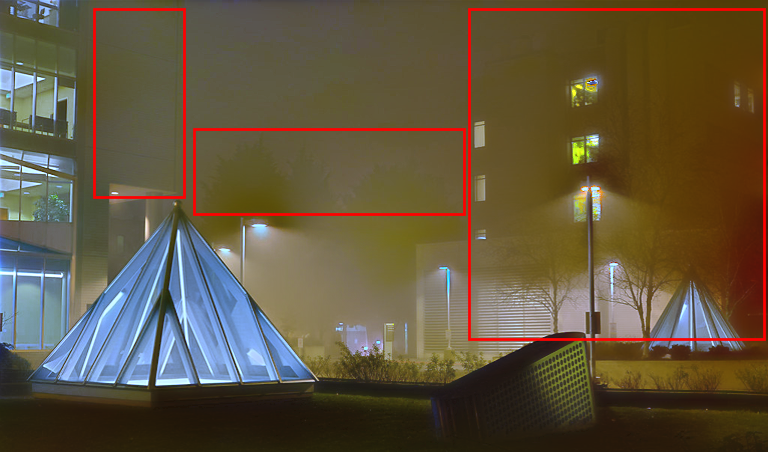}}\hfill
	\subfloat[Wang-22~\cite{wang2022variational}]{\includegraphics[width=0.245\textwidth]{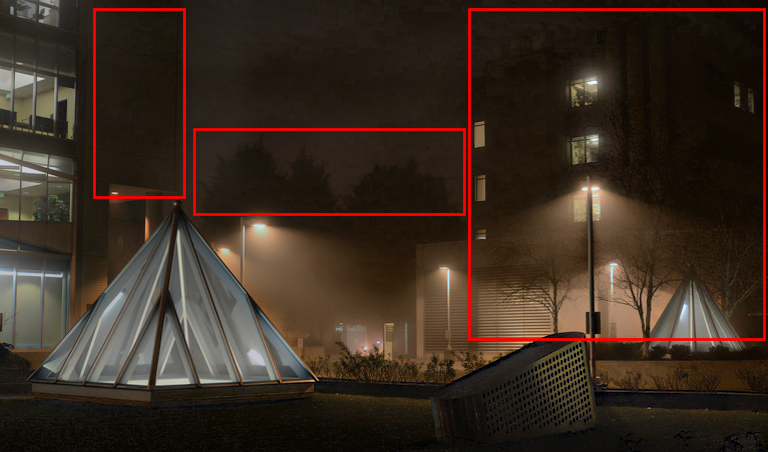}}\hfill
	\subfloat[Zhang~\cite{zhang2020nighttime}]{\includegraphics[width=0.245\textwidth]{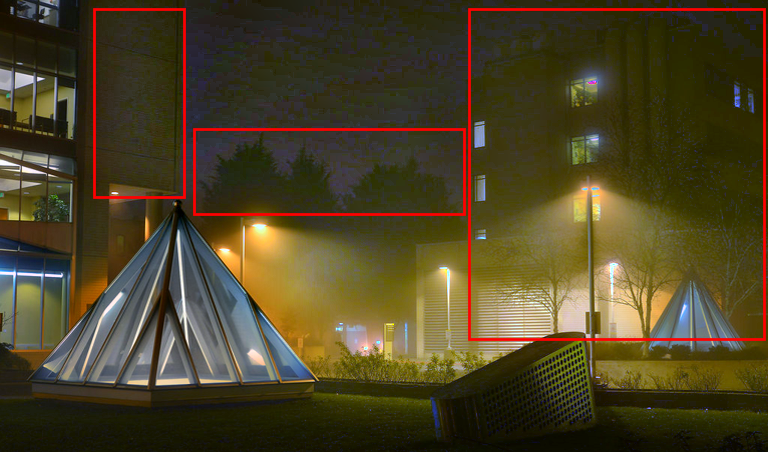}}\hfill
	\subfloat[Yan-20~\cite{yan2020nighttime}]{\includegraphics[width=0.245\textwidth]{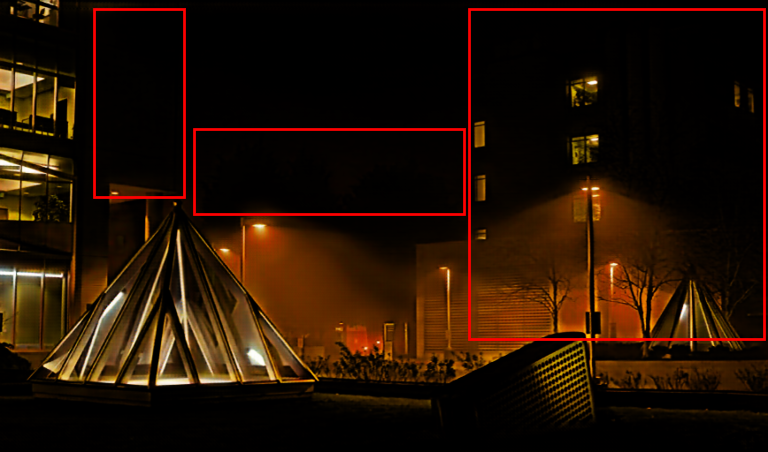}}\hfill
	\subfloat[Yu-19~\cite{yu2019nighttime}]{\includegraphics[width=0.245\textwidth]{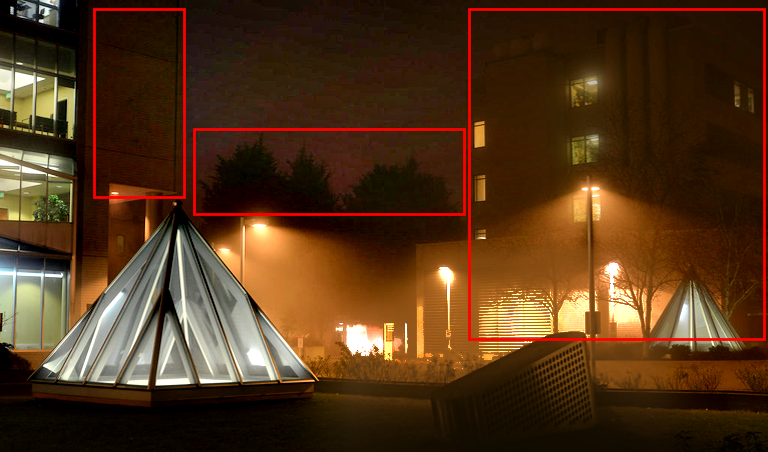}}\hfill
	\subfloat[Zhang~\cite{zhang2017fast}]{\includegraphics[width=0.245\textwidth]{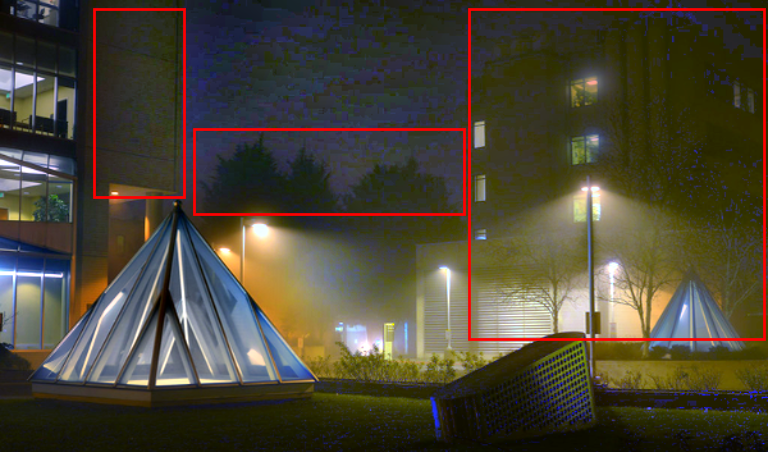}}\hfill
	\vspace{0.03in}
	\setcounter{subfigure}{0}
	\subfloat[Input]{\includegraphics[width=0.245\textwidth]{figures/Input/038.png}}\hfill
	\subfloat[Ours]{\includegraphics[width=0.245\textwidth]{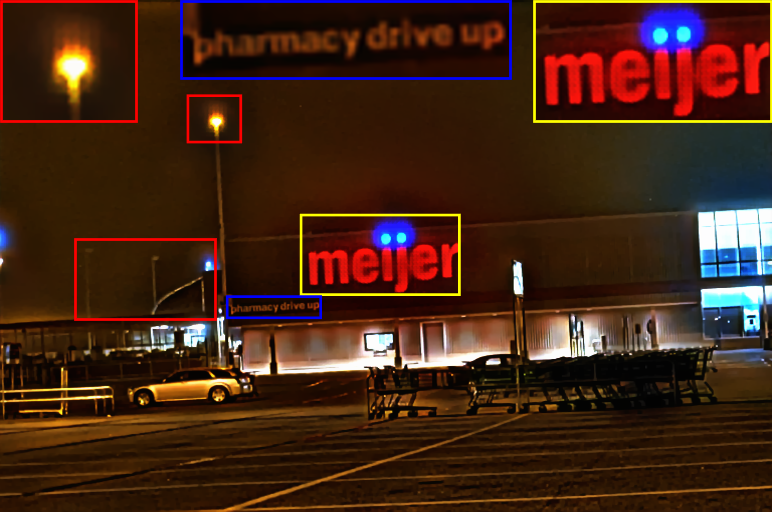}}\hfill
	\subfloat[Liu-22~\cite{liu2022nighttime}]{\includegraphics[width=0.245\textwidth]{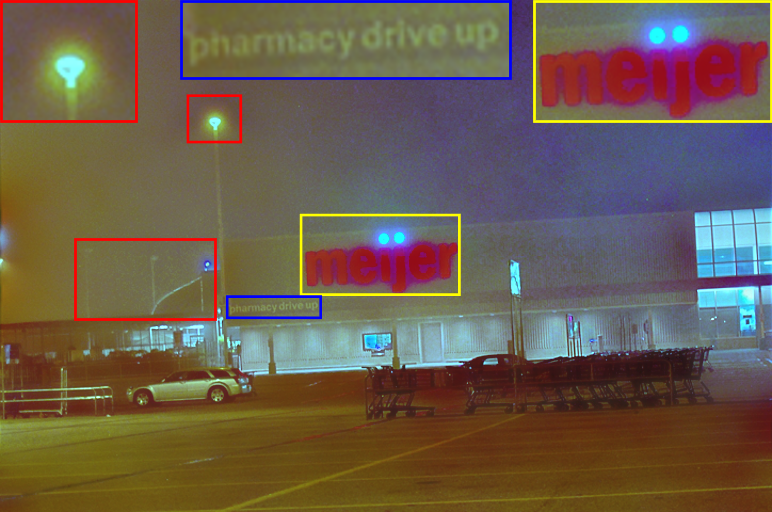}}\hfill
	\subfloat[Wang-22~\cite{wang2022variational}]{\includegraphics[width=0.245\textwidth]{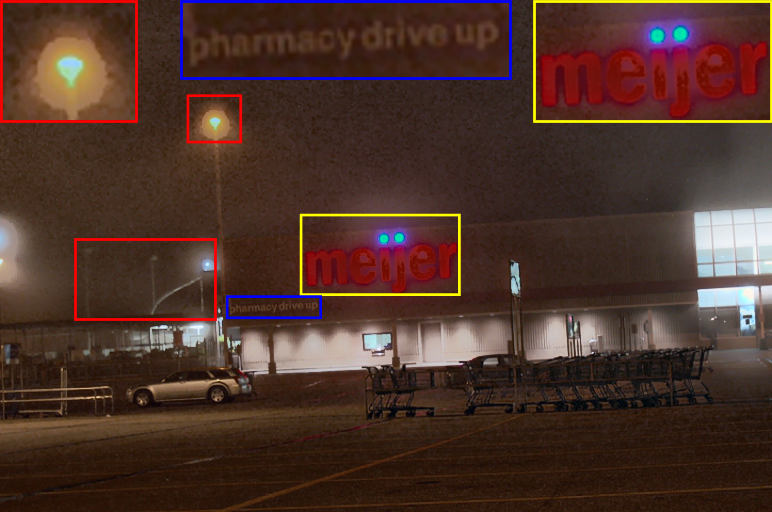}}\hfill
	\subfloat[Zhang~\cite{zhang2020nighttime}]{\includegraphics[width=0.245\textwidth]{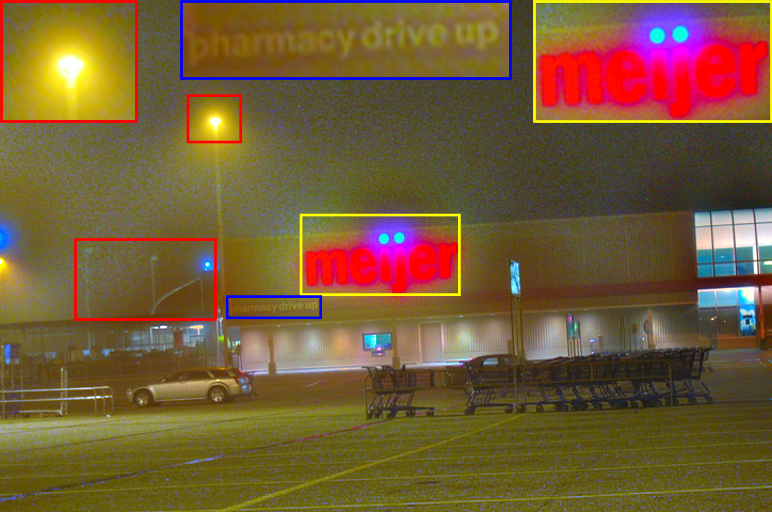}}\hfill	
	\subfloat[Yan-20~\cite{yan2020nighttime}]{\includegraphics[width=0.245\textwidth]{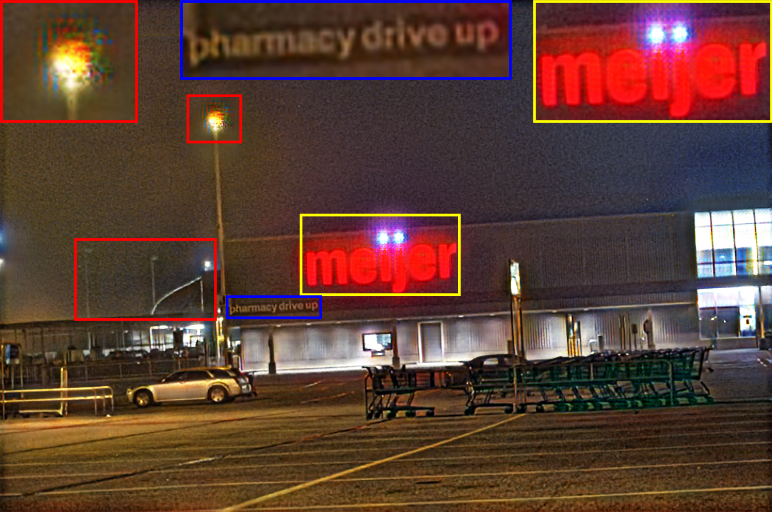}}\hfill
	\subfloat[Yu-19~\cite{yu2019nighttime}]{\includegraphics[width=0.245\textwidth]{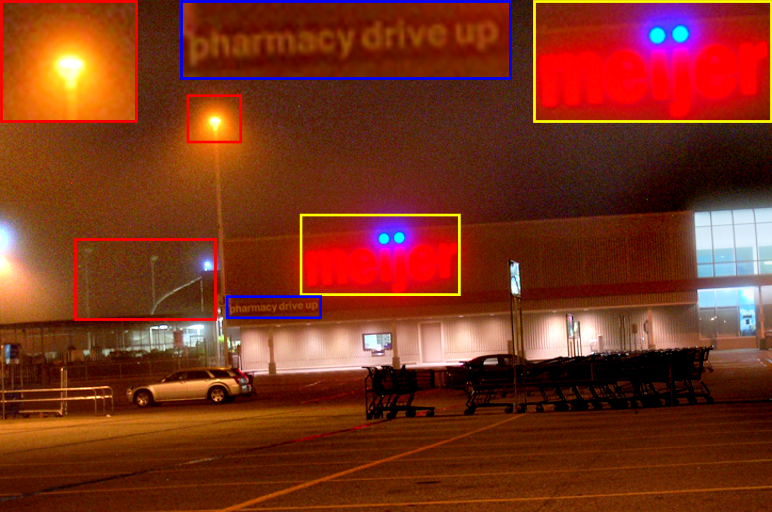}}\hfill		
	\subfloat[Ancuti-20~\cite{ancuti2020day}]{\includegraphics[width=0.245\textwidth]{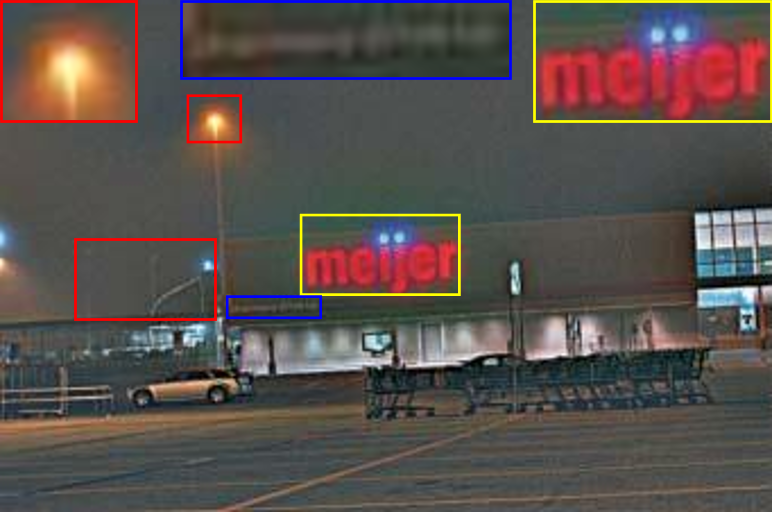}}\hfill
	\caption{Visual comparisons of different nighttime dehazing methods on real nighttime hazy scenes.
		Our results are more realistic and effective in nighttime dehazing. Zoom-in for better visualization.}
	\label{fig:1}
\end{figure*}

\section{Experimental Results}
\label{sec:experiments}
\subsection{Datasets}

\noindent \textbf{GTA5}~\cite{yan2020nighttime} is a synthetic nighttime dehazing dataset, which is generated by the GTA5 game engine. It includes 864 paired images, where 787 paired images are used as the training set and the rest images are taken as the test set.

\noindent \textbf{RealNightHaze} is a real-world night dehazing dataset. It includes 440 night hazy images, where 150 images are from \cite{zhang2020nighttime}, 200 images are from \cite{yan2020nighttime} and the rest images are collected from the Internet.

\subsection{Comparison on Synthetic Datasets}
In this section, we compare our method with existing state-of-the art methods, including Yan~\cite{yan2020nighttime}, Zhang~\cite{zhang2014nighttime}, Li~\cite{li2015nighttime} Ancuti~\cite{ancuti2016night}, Zhang~\cite{zhang2017fast}, Yu~\cite{yu2019nighttime}, Zhang~\cite{zhang2020nighttime}, Liu~\cite{liu2023multi} and Wang~\cite{wang2022variational}. 
The summary of the main differences is show in~\Cref{tb:compare}.
The experimental results are shown in~\Cref{tb:tb_real_syn}. 
It can be observed that our method achieves a significant performance improvement.
We adopt two widely used metrics PSNR, SSIM in generation~\cite{wang2023seeing} and restoration~\cite{ye2022underwater,li2023learning,li2020learning,xiao2021space,xiao2020space,jin2021dc,xiao2023dive,chen2022snowformer,chen2023msp} tasks.
Our method achieves a PSNR of 30.383 and a SSIM of 0.904, outperforming Yan's method~\cite{yan2020nighttime} by 14$\%$ and 5$\%$, respectively. 
This is because our method learns from the APSF-guided glow rendering and thus effectively removes the glow effects. 
Another advantage is that we introduce a gradient-adaptive convolution to capture the edges and textures. 
The obtained edges and textures are then used to enhance the structural details of the enhanced images, leading to superior performance.

\subsection{Comparison on Real-World Datasets}
\Cref{fig:1} show the qualitative results, including Liu~\cite{liu2023multi}, Wang~\cite{wang2022variational}, Zhang~\cite{zhang2020nighttime}, Yan~\cite{yan2020nighttime}, Yu~\cite{yu2019nighttime}, Zhang~\cite{zhang2017fast} and our method. 
It can be found that our method significantly enhances the visibility of nighttime hazy images. Specifically, most state-of-the-art methods cannot sufficiently remove haze since their methods suffer from the domain gap between the synthetic datasets and real-world images. Yan et al~\cite{yan2020nighttime} proposes a semi-supervised framework for nighttime foggy removal and can remove most hazy effects. However, their method over-suppresses hazy images, and thus their outputs become too dark. 

In contrast, our method handles glow and low-light conditions. As shown in \Cref{fig:1} (b), our method not only removes the haze of the input images but also enhances the light. For instance, the details of trees and buildings are clear. This is because our method simulates the glow rendering by utilizing Atmospheric Point Spread Function (APSF) and thus can effectively remove haze or glow in a real-world night hazy image. Moreover, we propose a gradient-adaptive convolution to capture the edges and textures from hazy images. The captured edges and textures are then used to boost the details of images, leading to superior performance. Furthermore, we introduce an attention map to enhance the low-light regions. As a result, our method achieves a significant performance improvement.

We also conduct user studies on real-world night hazy images. The experiment results are shown in~\Cref{tb:tb_user}. It can be found that our methods get the highest scores in all aspects.

\subsection{Ablation Study}

Our framework includes three core parts: APSF-guided glow rendering, gradient-adaptive convolution and attention-guided enhancement.
To prove the effectiveness of each part, we conduct ablation studies on real-world night hazy images. 

\noindent \textbf{APSF-guided glow rendering} \Cref{fig:apsf} (bottom) shows the results of our glow rendering. 
We can obverse that our method can accurately detect the location of light sources (middle). Also, the rendered results effectively simulate the glow effects. 

\begin{figure}[t]
	\captionsetup[subfloat]{farskip=1pt}
	\setcounter{subfigure}{0}
	\subfloat[Input $I_h$]{\includegraphics[width=0.196\columnwidth,height=0.2\columnwidth]{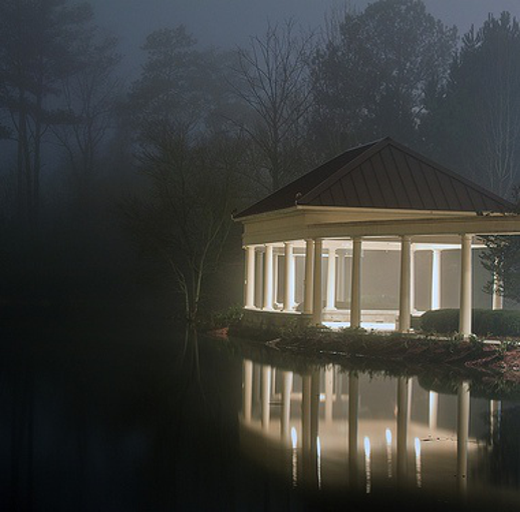}}\hfill
	\subfloat[w/o $\mathcal{L}_g$]{\includegraphics[width=0.196\columnwidth,height=0.2\columnwidth]{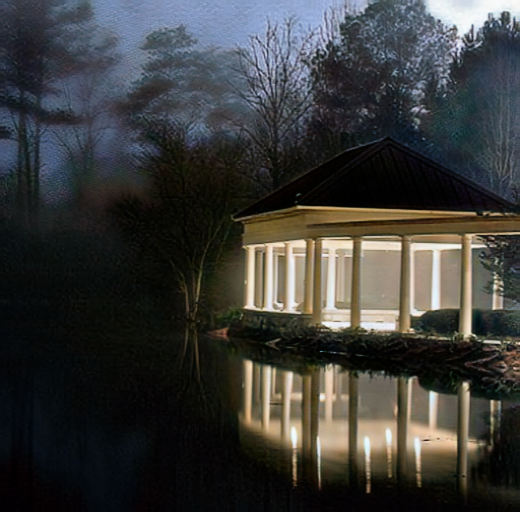}}\hfill
	\subfloat[w/o $\mathcal{L}_k$]{\includegraphics[width=0.196\columnwidth,height=0.2\columnwidth]{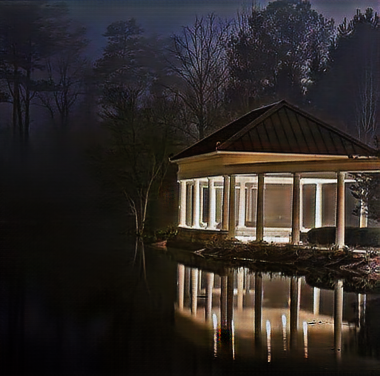}}\hfill
	\subfloat[w/ GAC]{\includegraphics[width=0.196\columnwidth,height=0.2\columnwidth]{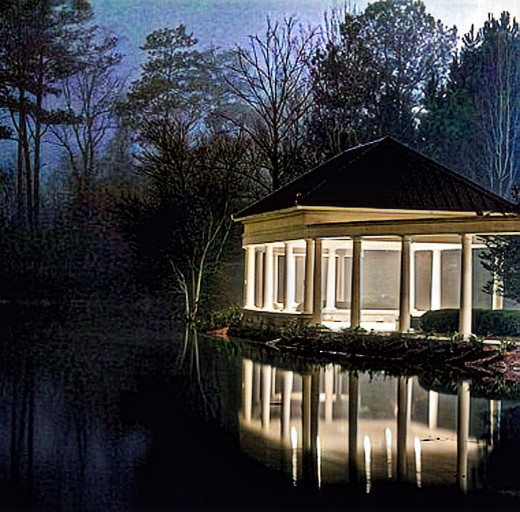}}\hfill
	\subfloat[Enhance]{\includegraphics[width=0.196\columnwidth,height=0.2\columnwidth]{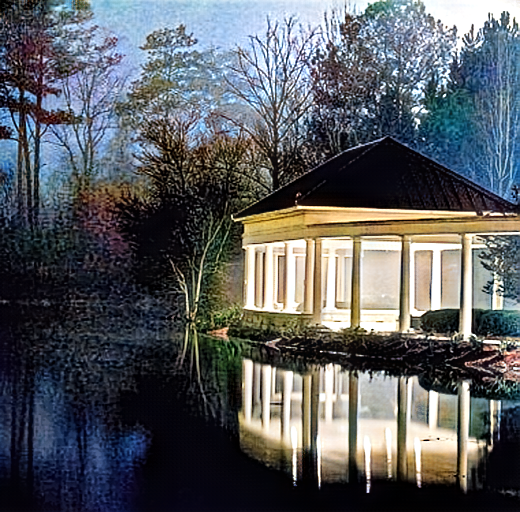}}\hfill
	\caption{Ablation study of gradient adaptive conv. (GAC).}
	\label{fig:ab_gac}
\end{figure}

\noindent \textbf{Gradient-adaptive convolution} \Cref{fig:edge_texture} show the results of the gradient edge maps (middle) and textures (bottom). 
\Cref{fig:ab_gac} (b) and (c) show results without gradient loss $\mathcal{L}_g$ and without kernel loss $\mathcal{L}_k$, and (d) is with gradient capture convolution.
It can be found that our gradient capture convolution can effectively preserve the structural details of hazy images. 
By taking full advantage of the gradient maps and textures, our framework generates sharper results. 

\begin{figure}[t]
	\captionsetup[subfigure]{font=small, labelformat=empty}
	\setcounter{subfigure}{0}
	\subfloat[Input $I_h$]{\includegraphics[width=0.245\columnwidth,height=0.25\columnwidth]{figures/Input/flickr3_real_B.png}}\hfill
	\subfloat[w/o Enhance $O_c$]{\includegraphics[width=0.245\columnwidth,height=0.25\columnwidth]{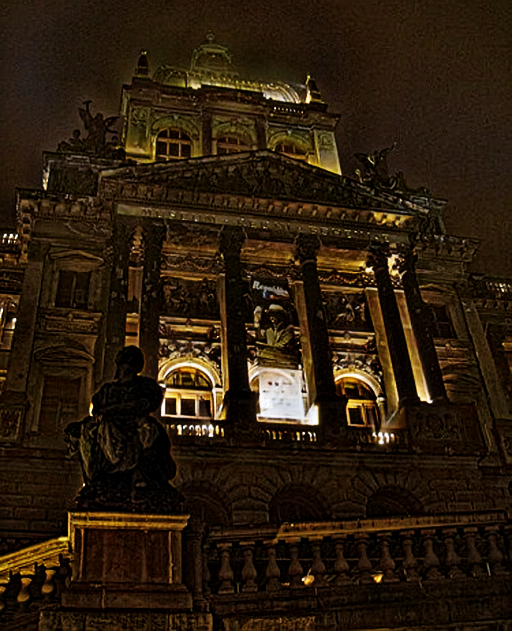}}\hfill
	\subfloat[Attention $A$]{\includegraphics[width=0.245\columnwidth,height=0.25\columnwidth]{figures/map/flickr3_real_B.png}}\hfill
	\subfloat[w/ Enhance $O_e$]{\includegraphics[width=0.245\columnwidth,height=0.25\columnwidth]{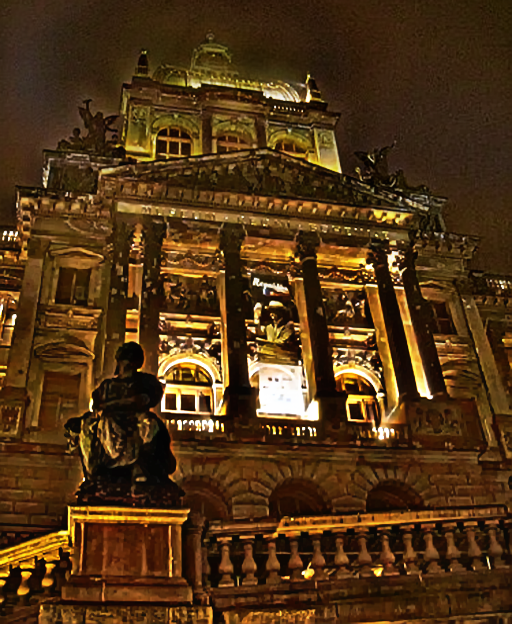}}\hfill\\
	\caption{Ablation study of low-light enhancement.}
	\label{fig:ab}
\end{figure}

\noindent \textbf{Low Light Enhancement} \Cref{fig:ab} shows the results of low-light enhancement from (b) to (d).

\section{Conclusion}
\label{sec:conclusion}
In this paper, we have proposed a novel nighttime visibility enhancement framework, addressing both glow and low-light conditions. Our framework includes three core ideas: APSF-guided glow rendering, gradient-adaptive convolution and attention-guided low-light enhancement. 
Our framework suppresses glow effects via learning from the APSF-guided glow rendering data.
Thanks to our APSF-guided glow rendering, we allow to use a semi-supervised method to optimize our network, thus handling glow effects in different light sources.
Our gradient-adaptive convolution is proposed to capture edges or textures from a nighttime hazy image. 
Benefiting from the captured edges or textures, our framework effectively preserves the structural details.
Our low-light region enhancement boosts the intensity of dark or over-suppressed regions via attention map.
Both quantitative and qualitative experiments show that our method achieves a significant performance improvement. 
Moreover, the ablation study proves the effectiveness of each core idea.
Handling scenarios with diverse domain shifts~\cite{zhang2023adaptive} will be a focus of our future research.

\begin{acks}
This research/project is supported by the National Research Foundation, Singapore under its AI Singapore Programme (AISG Award No: AISG2-PhD/2022-01-037[T]).
\end{acks}

\bibliographystyle{ACM-Reference-Format}
\balance
\bibliography{egbib}

\appendix

\begin{table*}
	\renewcommand\arraystretch{1.1}
	\caption{The comparison with dataset~\cite{zhang2020nighttime}, for PSNR and SSIM, higher values indicate better performance. We tested our method and Yan~\cite{yan2020nighttime} on a GPU GTX 3090 using an image resolution of 512x512 to measure the runtime. Other numbers in the table are borrowed from~\cite{zhang2020nighttime}.}
	\begin{tabular}{c|c|c|cc|cc|cc|c|c}\hline
		\multirow{2}{*}{Type} & \multirow{2}{*}{Method} & \multirow{2}{*}{Venue}  & \multicolumn{2}{c|}{NHR} & \multicolumn{2}{c|}{NHM} & \multicolumn{2}{c|}{NHC} & \multirow{2}{*}{Parameters} & \multirow{2}{*}{Time (s)} \\\cline{4-9}
		&  &  & PSNR $\uparrow$ & SSIM $\uparrow$  & PSNR $\uparrow$ & SSIM $\uparrow$ & PSNR $\uparrow$  & SSIM $\uparrow$ & & \\\hline
		& Zhang NDIM~\cite{zhang2014nighttime}& \textit{ICIP'14} &14.31 & 0.53  & 14.58 & 0.56 & 11.12 &  0.29 &- &5.63\\
		& Li GS~\cite{li2015nighttime} & \textit{ICCV'15}      &17.32 & 0.63  & 16.84 & 0.69 & 18.84 & 0.55 &- &22.52 \\
		Opti.& Zhang FAST-MRP~\cite{zhang2017fast}& \textit{CVPR'17}    &16.95 & 0.67  &13.85 & 0.61 & 19.17 & 0.58  &- &0.236\\
		& Zhang MRP~\cite{zhang2017fast} & \textit{CVPR'17}&19.93 &0.78 & 17.74 & 0.71 & 23.02 & 0.69  &- &1.769\\
		& Zhang OSFD~\cite{zhang2020nighttime}&  \textit{MM'20} & 21.32 & 0.80 & 19.75 & 0.76 &23.10 &0.74  &- &0.576\\
		& Zhang NDNET~\cite{zhang2020nighttime}&  \textit{MM'20} & \bf{28.74} & \bf{0.95} & 21.55 &0.91 &26.12 &0.85 &- &\bf{0.0074} \\\hline
		Learning & Yan~\cite{yan2020nighttime} & \textit{ECCV'20} &21.05 &0.62 &17.54&0.45 & 15.06 & 0.46 &50M &0.97\\\hline
		Learning & Ours &  \textit{MM'23} & 26.56 &0.89 & \bf{33.76}& \bf{0.92} & \bf{38.86} & \bf{0.97} &\bf{21M} &1.20\\\hline
	\end{tabular}
	\label{table:comparisons}
\end{table*}

\vspace{-0.3cm}
\begin{figure*}
	\centering
	\captionsetup[subfloat]{farskip=2pt}
	\setcounter{subfigure}{0}
	\subfloat{\includegraphics[width=0.245\textwidth]{figures/APSF/clean/000355.png}}\hfill
	\subfloat{\includegraphics[width=0.245\textwidth]{figures/APSF/clean/000159.png}}\hfill
	\subfloat{\includegraphics[width=0.245\textwidth]{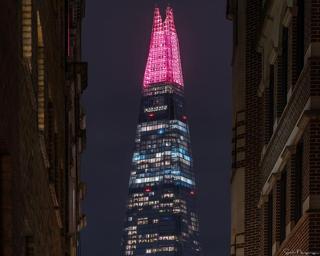}}\hfill
	\subfloat{\includegraphics[width=0.245\textwidth]{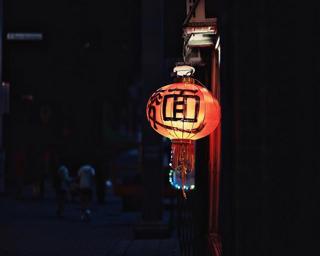}}\hfill
	\setcounter{subfigure}{0}
	\subfloat{\includegraphics[width=0.245\textwidth]{figures/APSF/degrad/000355.png}}\hfill
	\subfloat{\includegraphics[width=0.245\textwidth]{figures/APSF/degrad/000159.png}}\hfill
	\subfloat{\includegraphics[width=0.245\textwidth]{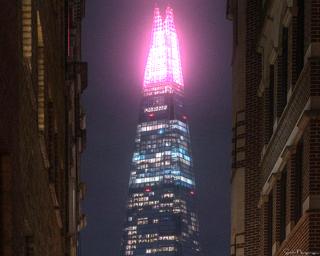}}\hfill
	\subfloat{\includegraphics[width=0.245\textwidth]{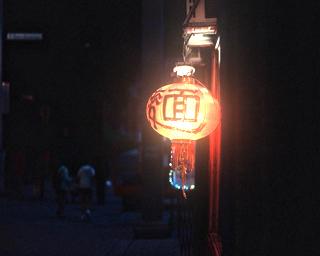}}\hfill
	\vspace{0.1cm}
	\caption{We show with APSF, we can render glow $I_g$ (bottom) on night clean $I_c$ (top).}
	\label{fig:apsf1}
\end{figure*}

\vspace{-0.3cm}
\begin{figure*}
	\centering
	\captionsetup[subfloat]{farskip=2pt}
	\setcounter{subfigure}{0}
	\subfloat{\includegraphics[width=0.245\textwidth]{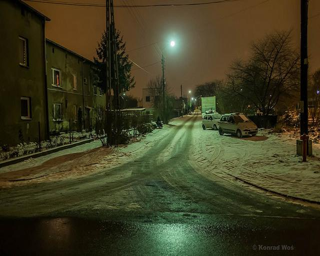}}\hfill
	\subfloat{\includegraphics[width=0.245\textwidth]{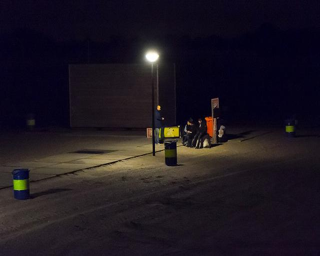}}\hfill
	\subfloat{\includegraphics[width=0.245\textwidth]{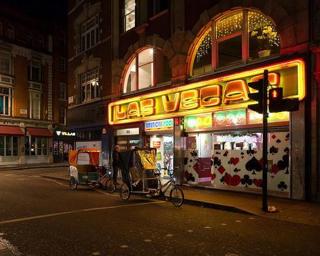}}\hfill
	\subfloat{\includegraphics[width=0.245\textwidth]{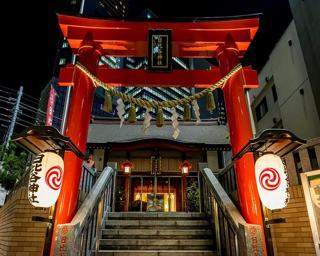}}\hfill
	\setcounter{subfigure}{0}
	\subfloat{\includegraphics[width=0.245\textwidth]{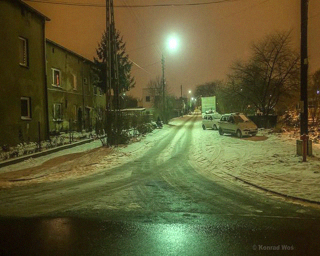}}\hfill
	\subfloat{\includegraphics[width=0.245\textwidth]{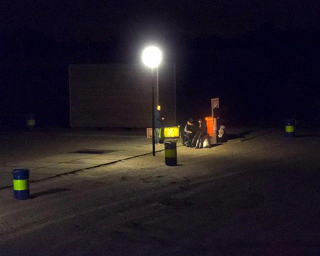}}\hfill
	\subfloat{\includegraphics[width=0.245\textwidth]{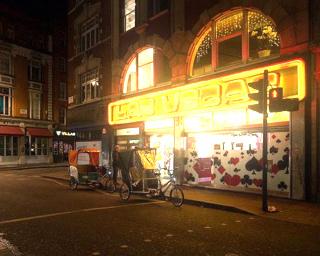}}\hfill
	\subfloat{\includegraphics[width=0.245\textwidth]{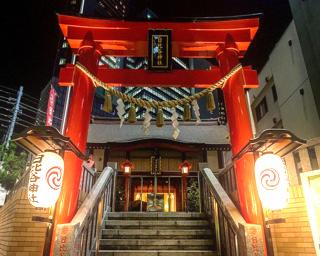}}\hfill
	\vspace{0.1cm}
	\caption{We show with APSF, we can render glow $I_g$ (bottom) on night clean $I_c$ (top).}
	\label{fig:apsf2}
\end{figure*}

\vspace{-0.3cm}
\begin{figure*}
	\centering
	\captionsetup[subfloat]{farskip=2pt}
	\setcounter{subfigure}{0}
	\subfloat{\includegraphics[width=0.245\textwidth]{figures/APSF/clean/000008.png}}\hfill
	\subfloat{\includegraphics[width=0.245\textwidth]{figures/APSF/clean/000174.png}}\hfill
	\subfloat{\includegraphics[width=0.245\textwidth]{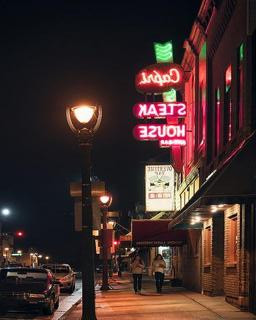}}\hfill
	\subfloat{\includegraphics[width=0.245\textwidth]{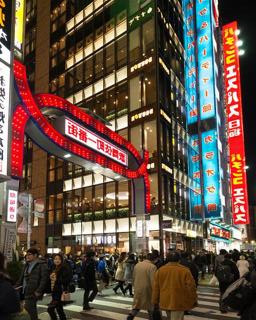}}\hfill
	\setcounter{subfigure}{0}
	\subfloat{\includegraphics[width=0.245\textwidth]{figures/APSF/degrad/000008.png}}\hfill
	\subfloat{\includegraphics[width=0.245\textwidth]{figures/APSF/degrad/000174.png}}\hfill
	\subfloat{\includegraphics[width=0.245\textwidth]{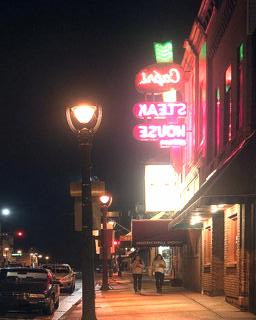}}\hfill
	\subfloat{\includegraphics[width=0.245\textwidth]{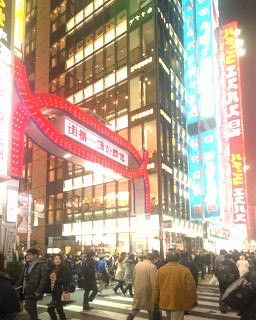}}\hfill
	\vspace{0.1cm}
	\caption{We show with APSF, we can render glow $I_g$ (bottom) on night clean $I_c$ (top).}
	\label{fig:apsf3}
\end{figure*}

\vspace{-0.3cm}
\begin{figure*}
	\centering
	\captionsetup[subfloat]{farskip=2pt}
	\setcounter{subfigure}{0}
	\subfloat{\includegraphics[width=0.245\textwidth]{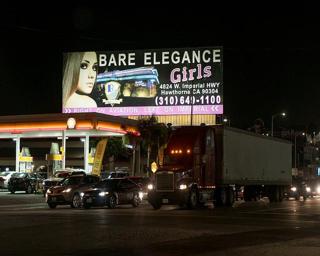}}\hfill
	\subfloat{\includegraphics[width=0.245\textwidth]{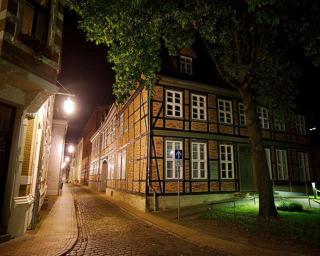}}\hfill
	\subfloat{\includegraphics[width=0.245\textwidth]{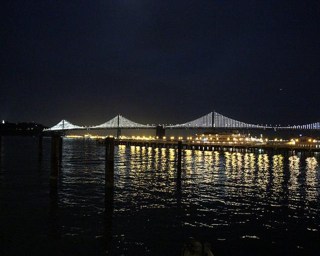}}\hfill
	\subfloat{\includegraphics[width=0.245\textwidth]{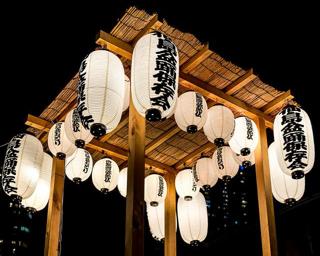}}\hfill
	\setcounter{subfigure}{0}
	\subfloat{\includegraphics[width=0.245\textwidth]{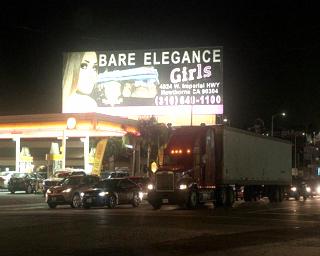}}\hfill
	\subfloat{\includegraphics[width=0.245\textwidth]{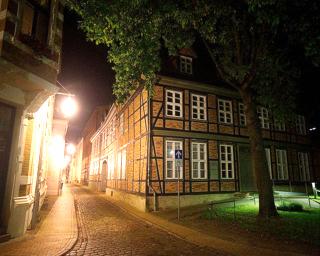}}\hfill
	\subfloat{\includegraphics[width=0.245\textwidth]{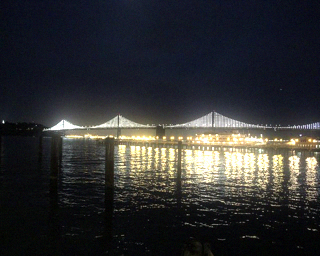}}\hfill
	\subfloat{\includegraphics[width=0.245\textwidth]{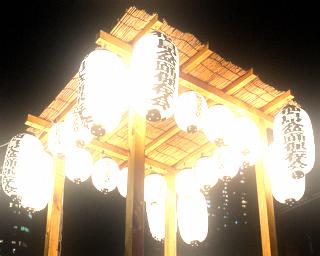}}\hfill
	\vspace{0.1cm}
	\caption{We show with APSF, we can render glow $I_g$ (bottom) on night clean $I_c$ (top).}
	\label{fig:apsf4}
\end{figure*}

\end{document}